%% file: paper.tex
\documentclass[10pt,journal,compsoc]{IEEEtran}
\ifCLASSOPTIONcompsoc
  \usepackage[nocompress]{cite}
\else
  \usepackage{cite}
\fi
\ifCLASSINFOpdf
  \usepackage[pdftex]{graphicx}
  \graphicspath{{../figs/}}
  \DeclareGraphicsExtensions{.pdf,.jpeg,.png}
\else
\fi

\usepackage{stfloats}
\usepackage{amsmath}
\usepackage{amssymb}
\usepackage{xr}
\usepackage{dsfont}
\usepackage{pgfplots}
\usepackage{xcolor}
\usepackage{lipsum}
\usepackage{graphicx}
\usepackage{caption}
\usepackage{subcaption}
\usepackage{placeins}
\usepackage[hidelinks]{hyperref}
\usepackage{ragged2e}
\usepackage{framed}
\usepackage{soul}
\usepackage{tabu}
\usepackage{booktabs} 
\usepackage{subcaption}
\usepackage{hyperref}
\usepackage{times,epsfig}

\graphicspath{ {images/} }
\pgfplotsset{table/search path={data}}
\pgfplotsset{compat=1.13}
\usetikzlibrary{arrows,positioning,decorations,shapes,calc}
\usepgfplotslibrary{statistics}

\newcommand{\latinphrase}[1]{\textit{#1}}
\newcommand{\etal}{\latinphrase{et~al. }}

\usepackage{color}
\definecolor{darkgreen}{RGB}{47,109,79}
\definecolor{darkblue}{RGB}{57,79,99}
\definecolor{rosso}{RGB}{220,57,18}
\definecolor{giallo}{RGB}{255,153,0}
\definecolor{blu}{RGB}{102,140,217}
\definecolor{verde}{RGB}{16,150,24}
\definecolor{viola}{RGB}{153,0,153}
\definecolor{awesome}{rgb}{1.0, 0.13, 0.32}
\definecolor{ref}{rgb}{0.65,0.65,0.65} 
\usetikzlibrary{positioning,shadows}
\usetikzlibrary{patterns}

\colorlet{shadecolor}{yellow!40}

\usetikzlibrary{automata,positioning}

\usepackage{todonotes}

\newcommand{\BlackBox}{\rule{1.5ex}{1.5ex}}  

\newcommand{\descr}[1]{\smallskip\noindent\textbf{#1}}

\begin{document}
	
	\makeatletter
	\pgfplotsset{
		boxplot/hide outliers/.code={
			\def\pgfplotsplothandlerboxplot@outlier{}%
		}
	}
	\makeatother
	
%
\title{A Hybrid Deep Learning Architecture for Privacy-Preserving Mobile Analytics}
%
%
%
%

\author{Seyed~Ali~Osia,~Ali~Shahin~Shamsabadi,~Sina~Sajadmanesh,~Ali~Taheri, Kleomenis~Katevas,~Hamid~R.~Rabiee,~Nicholas D. Lane,~Hamed~Haddadi
\IEEEcompsocitemizethanks{\IEEEcompsocthanksitem Seyed~Ali~Osia, Ali~Taheri and Hamid~R.~Rabiee are with the Department of Computer Engineering, Sharif University of Technology, Iran.
\IEEEcompsocthanksitem Ali~Shahin~Shamsabadi is with the School of Electronic Engineering and Computer Science, Queen Mary University of London. 
\IEEEcompsocthanksitem Sina~Sajadmanesh is with Idiap Research Institute and Ecole Polytechnique Federale de Lausanne (EPFL).
\IEEEcompsocthanksitem Hamed~Haddadi and Kleomenis~Katevas are with the Faculty of Engineering, Imperial College London.
\IEEEcompsocthanksitem Nicholas~D.~Lane is with the Samsung AI Center \& University of Oxford.}
}

\IEEEtitleabstractindextext{%
\input{abstract.tex}

}

\maketitle

\IEEEdisplaynontitleabstractindextext

%
\IEEEpeerreviewmaketitle

\input{introduction}

\input{decomposition}
\input{theory}
\input{evaluation}
\input{experiments}
\input{related}

\input{conclusions}

%

\ifCLASSOPTIONcompsoc
\else
\fi



\ifCLASSOPTIONcaptionsoff
  \newpage
\fi



\bibliographystyle{IEEEtran}
\bibliography{IEEEabrv,ref2}


%

%
\vspace*{-4\baselineskip}
\begin{IEEEbiography}[{\includegraphics[height=1in,keepaspectratio]{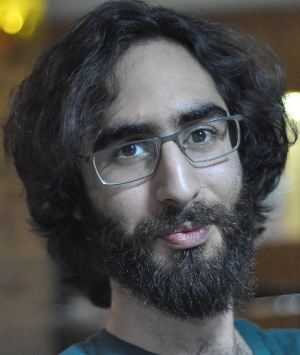}}]{Seyed Ali Osia}
	received his B.Sc. degree in Software Engineering and his Ph.D. degree in Artificial Intelligence from Sharif University of Technology in 2014 and 2019, respectively. His research interests include Statistical Machine Learning, Deep Learning, Privacy and Computer Vision. 
\end{IEEEbiography}
\vspace*{-5\baselineskip}
\begin{IEEEbiography}[{\includegraphics[height=1in,clip,keepaspectratio]{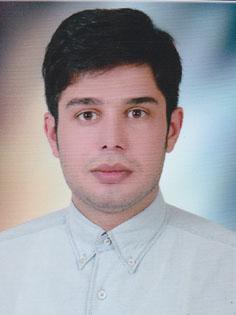}}]{Ali Shahin Shamsabadi}
	received his B.S. degree in electrical engineering from Shiraz University of Technology, in 2014, and the M.Sc. degree in electrical engineering (digital) from the Sharif University of Technology, in 2016. Currently, he is a Ph.D. candidate at  Queen Mary University of London. His research interests include deep learning and data privacy protection in distributed and centralized learning. 
\end{IEEEbiography}
\vspace*{-5\baselineskip}
\begin{IEEEbiography}[{\includegraphics[height=1in,clip,keepaspectratio]{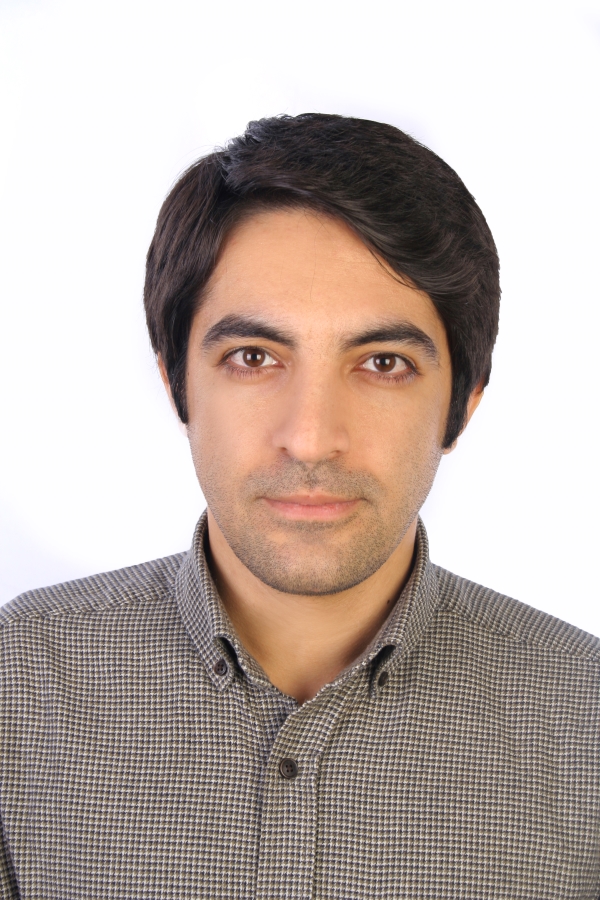}}]{Sina~Sajadmanesh}
	received his B.Sc. degree in Computer Engineering from the University of Isfahan in 2014, and his M.Sc. degree in Information Technology Engineering from Sharif University of Technology in 2016. He is currently a Ph.D. student in the School of Engineering at EPFL and a research assistant in Idiap Research Institute. He is interested in Machine Learning, Data Mining, and Social Computing.
\end{IEEEbiography}
\vspace*{-5\baselineskip}
\begin{IEEEbiography}[{\includegraphics[height=1in,clip,keepaspectratio]{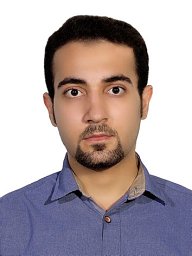}}]{Ali Taheri}
	received his B.Sc. degree in Software Engineering from Shahid Beheshti University in 2015. He received his M.Sc. degree in Artificial Intelligence from Sharif University of Technology in 2017. He is currently a Ph.D. student of Artificial Intelligence in the Department of Computer Engineering at Sharif University of Technology. His research interests include Deep Learning and Privacy.
\end{IEEEbiography}
\vspace*{-5\baselineskip}
\begin{IEEEbiography}[{\includegraphics[height=1in,clip,keepaspectratio]{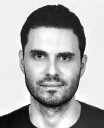}}]{Kleomenis~Katevas}
	received his B.Sc. degree in Informatics Engineering from the University of Applied Sciences of Thessaloniki in 2006, and his M.Sc. and Ph.D. degrees in Software Engineering from Queen Mary University of London in 2010 and 2018. He is currently a postdoctoral researcher at Imperial College London. His research interests include Mobile \& Ubiquitous Computing, Applied Machine Learning, Crowd Sensing and Human-Computer Interaction.
\end{IEEEbiography}
\vspace*{-4\baselineskip}
\begin{IEEEbiography}[{\includegraphics[height=1in,clip,keepaspectratio]{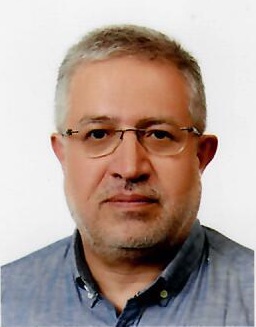}}]{Hamid R. Rabiee}(SM '07)
received his BS and MS degrees (with Great Distinction) in Electrical Engineering from CSULB, Long Beach, CA (1987, 1989), his EEE degree in Electrical and Computer Engineering from USC, Los Angeles, CA (1993), and his Ph.D. in Electrical and Computer Engineering from Purdue University, West Lafayette, IN, in 1996. From 1993 to 1996 he was a Member of Technical Staff at AT\&T Bell Laboratories. From 1996 to 1999 he worked as a Senior Software Engineer at Intel Corporation. He was also with PSU, OGI and OSU universities as an adjunct professor of Electrical and Computer Engineering from 1996-2000. Since September 2000, he has joined Sharif University of Technology, Tehran, Iran. He was also a visiting professor at the Imperial College of London for the 2017-2018 academic year. He is the founder of Sharif University Advanced Information and Communication Technology Research Institute (AICT), ICT Innovation Center, Advanced Technologies Incubator (SATI), Digital Media Laboratory (DML), Mobile Value Added Services Laboratory (VASL), Bioinformatics and Computational Biology Laboratory (BCB) and Cognitive Neuroengineering Research Center. He is also a consultant and member of AI in Health Expert Group at WHO. He has been the founder of many successful High-Tech start-up companies in the field of ICT as an entrepreneur. He is currently a Professor of Computer Engineering at Sharif University of Technology, and Director of AICT, DML, and VASL. He has received numerous awards and honors for his Industrial, scientific and academic contributions, and holds three patents. His research interests include statistical machine learning, Bayesian statistics, data analytics and complex networks with applications in social networks, multimedia systems, cloud and IoT privacy, bioinformatics, and brain networks.
\end{IEEEbiography}
\vspace*{-3.5\baselineskip}
\begin{IEEEbiography}[{\includegraphics[height=1in,clip,keepaspectratio]{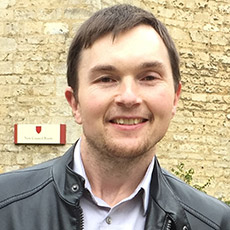}}]{Nicholas D. Lane}
	is an Associate Professor in the Computer Science department at the University of Oxford, where he leads the Machine Learning Systems lab (OxMLSys). Alongside his academic role, he is also a Program Director at the Samsung AI Center in Cambridge where his teams study on-device and distributed forms of machine learning. To find out more about his research please visit http://niclane.org and http://mlsys.cs.ox.ac.uk. 
\end{IEEEbiography}
\vspace*{-3.5\baselineskip}
\begin{IEEEbiography}[{\includegraphics[height=1in,clip,keepaspectratio]{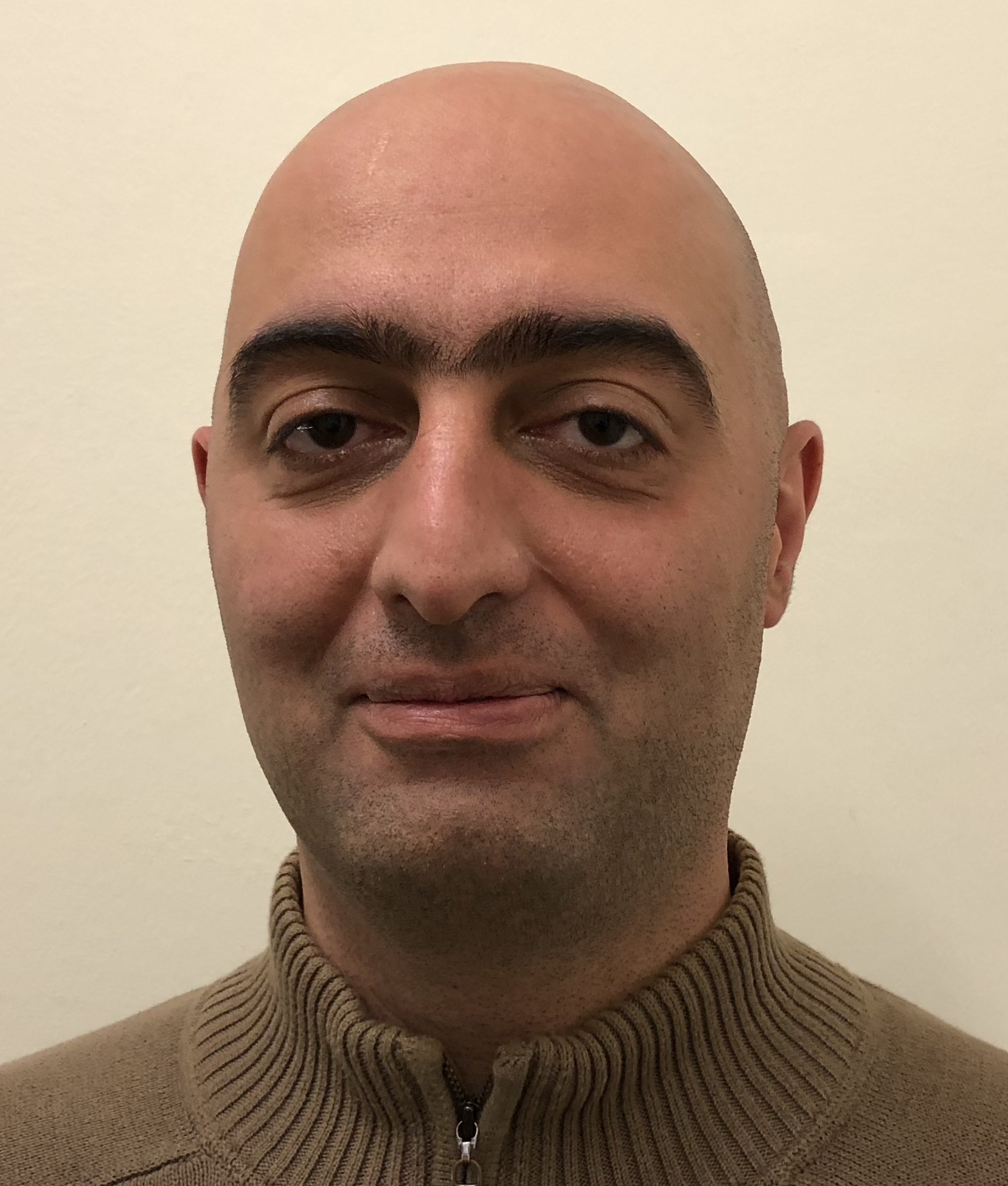}}]{Hamed Haddadi}
	received his B.Eng., M.Sc., and Ph.D. degrees from University College London. He was a postdoctoral researcher at Max Planck Institute for Software Systems in Germany, and a postdoctoral research fellow at Department of Pharmacology, University of Cambridge and The Royal Veterinary College, University of London, followed by few years as a Lecturer and consequently Senior Lecturer in Digital Media at Queen Mary University of London. He is currently a Senior Lecturer and the Deputy Director of Research in the Dyson School of Design Engineering at The Faculty of Engineering at Imperial College of London. He is interested in User-Centered Systems, IoT, Applied Machine Learning, and Data Security \& Privacy. 
\end{IEEEbiography}







\end{document}

%% file: abstract.tex
\begin{abstract}

%

{Internet of Things (IoT) devices and applications are being deployed in our homes and workplaces. These devices often rely on continuous data collection to feed machine learning models. However, this approach introduces several privacy and efficiency challenges, as the service operator can perform unwanted inferences on the available data.} Recently, advances in edge processing have paved the way for more efficient, and private, data processing at the source for simple tasks and lighter models, though they remain a challenge for larger, and more complicated models. In this paper, we present a hybrid approach for breaking down large, complex deep neural networks for cooperative, privacy-preserving analytics. To this end, instead of performing the whole operation on the cloud, we let an IoT device to run the initial layers of the neural network, and then send the output to the cloud to feed the remaining layers and produce the final result. {In order to ensure that the user's device contains no extra information except what is necessary for the main task and preventing any secondary inference on the data, we introduce Siamese fine-tuning.} We evaluate the privacy benefits of this approach based on the information exposed to the cloud service. We also assess the local inference cost of different layers on a modern handset. Our evaluations show that by using Siamese fine-tuning and at a small processing cost, we can greatly reduce the level of unnecessary, potentially sensitive information in the personal data, and thus achieving the desired trade-off between utility, privacy and performance.
\end{abstract}

%
%
%

%% file: introduction.tex
\section{Introduction} 
\label{intro}

The increasing availability of connected IoT devices such as smartphones and cameras has made them an essential and inseparable part of our daily lives. The majority of these devices collect various forms of data and transfer it to the cloud in order to benefit from cloud-based data mining services such as recommendation systems, targeted advertising, security surveillance, health monitoring, and urban planning. Many of these devices are subsidized, and their applications are free, relying on information harvested from the users' data. {This practice has several privacy concerns and resource impacts for the users~\cite{narseoIMC,dontkill,stergiou2018algorithms, psannis2018advanced}}. For example, a cloud-based IoT application that offers emotion detection service on the user's camera as its primary task, may use this data for other tasks such as occupancy analysis, face recognition or scene understanding, which may not be desired for the user, putting her privacy at risk. Preserving individuals' privacy versus providing detailed data analytics faces a dichotomy in this space. Cloud-based machine learning algorithms can provide beneficial services (e.g., video editing tools or health apps), but the redundant data collected from the users could be used for unwanted purposes (e.g., face recognition for targeted social advertising).

{
\descr{Motivation.}
While complete data offloading to a cloud provider can have immediate or future potential privacy risks \cite{pournajaf2016,haris2014privacy}, techniques relying on performing complete analytics at the user end (on-premise solution) or encryption-based methods also come with their resource limitations and user experience penalties (see Section~\ref{sec:related} for a detailed discussion).} Apart from resource considerations, an analytics service or an app provider might not be keen on sharing its valuable and highly tuned model. Therefore, it is not always possible to assume local processing is a viable solution even if the task duration, memory, and processing requirements are not crucial for the user, or the task can be performed when users are not actively using their devices (e.g., when the device is being charged overnight).

In this paper, we focus on achieving an optimization trade-off between resource-hungry, on-device analytics, versus privacy-invasive cloud-based services. We design and evaluate a hybrid architecture where the local device and the cloud system collaborate on running a complex neural network that has previously been trained and fine-tuned on the cloud. The proposed framework is based on the idea that the end-user does not need to upload her raw data to the cloud (which can put her privacy at risk), nor to hide all the information employing cryptographic methods (which can be resource-hungry or overly complicated for the end-user's device). Instead, it is sufficient to preserve the necessary information for the service provider's \emph{main task} and discard any other irrelevant information which could be used for unwanted inferences from personal data as much as possible. In this way, we can augment the local device to benefit from the cloud processing efficiency and effectively addressing the privacy concerns.

Fig~\ref{fig:framework} depicts an overview of the proposed hybrid framework, in which the user and the cloud collaborate to process the user's data privately and efficiently. Our work relies on the assumption that the service provider releases a feature extractor module that is publicly verifiable in terms of privacy. Using this feature extractor, instead of sending personal data, the user performs a minimalistic analysis and extracts an \emph{exclusive feature} from her data and sends it to the service provider for subsequent analysis. The exclusive feature is then processed in the cloud, and the result yields back to the user. The fundamental challenge in using this framework is the design of the feature extractor module that removes irrelevant information to achieve an acceptable trade-off among the privacy, utility, and scalability.

\descr{Overview \& Contributions.}
We begin by introducing the proposed hybrid framework, which addresses the problem of \emph{``user data privacy in interaction with cloud services''}. In this paper, we focus on those services that utilize deep neural networks (DNNs) aiming to solve \textit{classification} problems as their main task. By decoupling layers of a pre-trained DNN into two parts, we shift initial layers of the network that act as the feature extractor module to the user's device, while keeping the remaining layers on the cloud. We demonstrate that the proposed solution does not have the overhead of running the whole deep model on the user's device, and it can be used effectively to preserve the privacy of users by preventing exposure of irrelevant information to the cloud. 

In order to evaluate our hybrid privacy-preserving framework, we consider two different main tasks: gender classification on face images and activity recognition on mobile sensor data. First, we use activity recognition as the main task and consider gender recognition as the privacy-invasive secondary task. Then we set gender recognition as the main task, and we use face recognition, which tries to reveal the identity of individuals in the images, as a sensitive secondary task. We use Convolutional Neural Networks (CNNs) as one of the most widely used DNN architectures \cite{Rich:2016:TBA:2896338.2897734, Druzhkov2016, wan2014deep} to build accurate models for gender and activity recognition. Then we fine-tune these models with our suggested Siamese architecture to preserve the privacy of users' data while maintaining an acceptable level of accuracy. To verify the proposed feature extractor, measure its privacy, and evaluate the efficiency of the model in removing irrelevant information, we introduce a new measure for privacy measurement that is an extension of the zero-one loss for classification. Besides, we use \textit{transfer learning} \cite{yosinski2014}, which proves that secondary inferences are not feasible. To give more insight into the performance of our framework, we use \textit{deep visualization}, which tries to reconstruct the input image using the exclusive feature \cite{dosovitskiy2016}. We also implement the gender classification model on a modern smartphone device and compare the complete on-premise solution with our hybrid architecture. Even though we report the evaluation results for a typical smartphone device, this framework can be extended to other devices with limited memory and processing capabilities, such as Raspberry Pi and similar IoT devices. Our main contributions in this paper include:

\begin{itemize}
	\item Introducing a hybrid user-cloud framework for the privacy preservation problem which utilizes a novel feature extractor as its core component.
	\item Proposing a novel technique for building the feature extractor based on Siamese architecture that enables privacy at the point of offloading to the cloud.
	\item Proposing a new measure to evaluate the privacy and verify the feature extractor module.
	\item Evaluating the framework across two deep learning architectures for gender classification and activity recognition, based on the proposed privacy measure, transfer learning, and deep visualization.
\end{itemize}

\descr{Paper Organization.} The rest of this paper is organized as follows. In the next section, we present a thorough overview of the proposed hybrid architecture. In Section \ref{DLembedding}, we describe how to deploy a pre-trained model to build the feature extractor module using Siamese fine-tuning, dimensionality reduction and noise addition mechanisms. Next, we introduce different methodologies we employed to measure the privacy of the proposed framework in Section \ref{evaluation}.
Extensive evaluations and experimental results are described in Section \ref{experiments}. After reviewing related works in Section \ref{sec:related}, we finally conclude the paper in Section \ref{sec:conclusions}.

%% file: decomposition.tex
\label{decomposition}
\section{Hybrid User-Cloud Framework}


\begin{figure}[t]
	\begin{center}
		\includegraphics[width=.9\columnwidth]{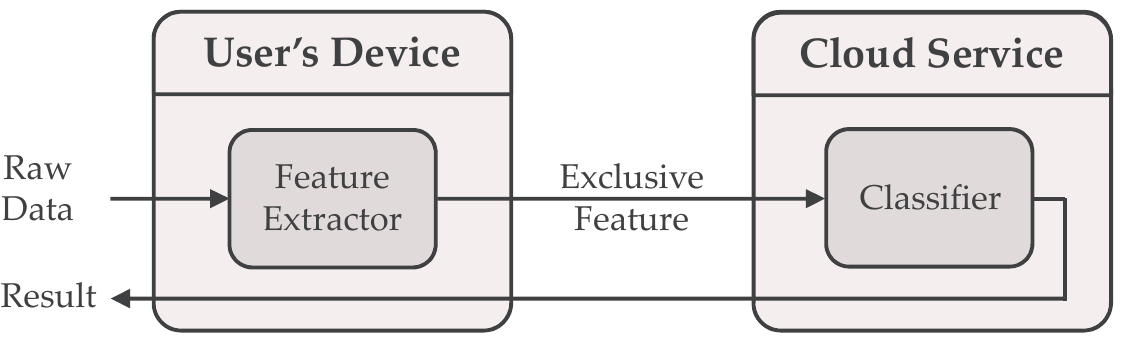}     
	\end{center}
	\caption{Hybrid privacy-preserving framework: The cloud service provides the feature extractor to the user's device, by which the device extracts an exclusive feature from the data and uploads it to the cloud for the rest of the process. The cloud runs a classifier on the exclusive feature and sends the result back to the user.}
	\label{fig:framework}
\end{figure}

In this section, we present our hybrid framework for cloud-based privacy-preserving analytics. Suppose the scenario in which the user would like to interact with a cloud-based service that makes use of a classification model for inferring a \emph{primary} measure of interest from the user's data as its main task. However, the end-user wants to prevent the exposure of her sensitive information to the service provider that could potentially put her privacy at risk. An example of this scenario can be seen when the end-user has an application on her phone, which sends sensor data to a cloud-based service provider for activity recognition (the main task) to react to her activities (the primary measure) differently (e.g., make the phone silent when sleeping). However, the end-user would not like the service provider to be able to infer other potentially sensitive information, such as her identity or gender, from her data. Note that in this case, the end-user does not choose to \emph{train} her classification model, but in fact, she wants to use a provided pre-trained cloud-based service in a safe, privacy-preserving manner.

While uploading the raw data to the cloud has the potential risk of revealing sensitive information, using cryptographic methods to encrypt personal data has its potential drawbacks, as discussed in Section \ref{sec:related}. Instead, an intermediary solution to this problem is to perform a minimal process on user's raw data in the client-side to extract a \emph{specially crafted feature}, which we refer to as \emph{exclusive feature}, and then upload this feature instead of the raw data to the cloud for further processing. In this case, the exclusive feature should have the following properties:
\begin{itemize}
	\item The exclusive feature must keep as much information relative to the \emph{primary measure} as possible to prevent the performance drop of the main task.
	\item It must hide or discard all the other unnecessary information to prevent the inference of any \emph{sensitive measure}.
\end{itemize}

As a result, according to Fig~\ref{fig:framework}, the user and the service provider collaborate in the following steps:

\begin{enumerate}
	\item The service provider determines how to extract the exclusive feature on the client-side by providing a \emph{feature extractor module} to the user.
	\item Using the provided feature extractor, the user extracts the exclusive feature from her data on the client-side and uploads it to the cloud for further processing.
	\item The service provider's \emph{classifier module} receives and processes the exclusive feature to yield and return the expected result to the user.
\end{enumerate}

Any classifier which can be separated to feature extractor and classier modules can be embedded in this framework. The main challenge of this work is to properly design the feature extractor module to output the exclusive feature constrained to keeping the \emph{primary} information, while discarding any other irrelevant information to the main task. Usually, these two objectives are contradictory because information removal may adversely affect the performance of the main task. Besides, due to the limitations of client-side processing, feature extraction needs to have minimal overhead on the user's device. Therefore, designing the feature extractor is the most challenging issue to address. 

{
In order to narrow down the scope of this paper, in the rest, we will focus on feed-forward neural networks (e.g., CNN or MLP) and address how to build and train the feature extractor module for them. By considering the accuracy of the classifier, we can guarantee the preservation of the primary information in the exclusive feature. The privacy preservation of the feature extractor can also be verified (either by the end-user or by a third party) using different methods, which will be discussed in Section~\ref{evaluation}. 
}

%% file: theory.tex
\label{theory}
\newcommand\norm[1]{\left\lVert#1\right\rVert}

\section{Deep Privacy Embedding}
\label{DLembedding}

\begin{figure}[t]
	\centering
	
	\includegraphics[width=\columnwidth]{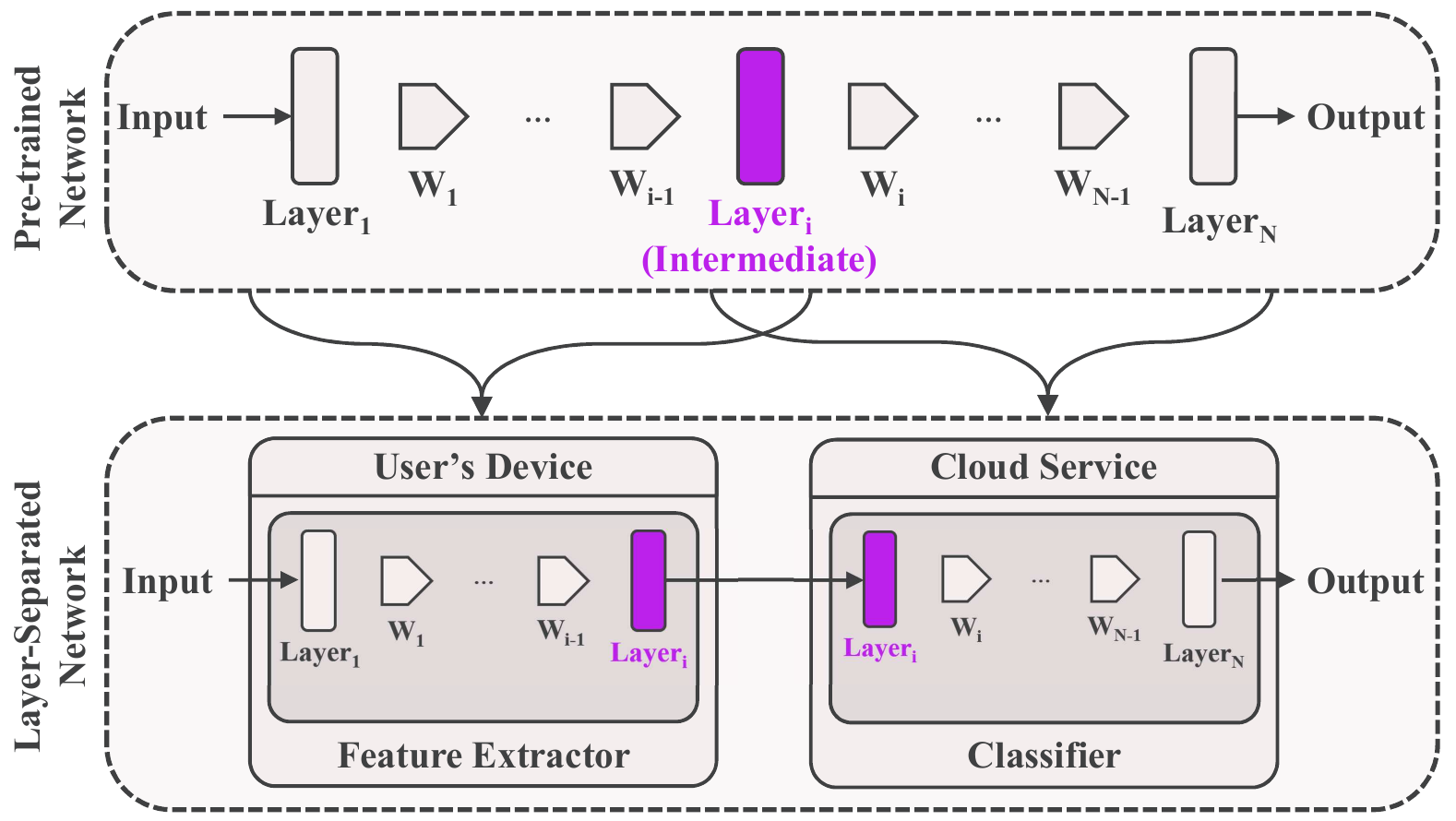}
	
	\caption{Simple Embedding of a DNN. The above network is a pre-trained DNN before applying layer separation, entirely resided in the cloud. After layer separation, the beginning part of the DNN forms the feature extractor and will be sent to the user's device for operation. The second part will remain and operate in the cloud.}
	\label{fig:simple}
\end{figure}

In this section, we address how to embed a pre-trained deep neural network (DNN), which classifies the primary measure (e.g., for gender recognition), into the proposed framework. We choose DNNs as the core learning method for our framework due to their increasing popularity in machine learning and data mining applications. Complex DNNs consist of many layers that can be embedded in this framework following the \emph{layer separation mechanism}. In this approach, we choose an \emph{intermediate layer} of the DNN as a pivot, based on which we split the whole network into two parts. The first part that forms the feature extractor module includes the input layer of the DNN up to the intermediate layer (inclusive). We refer to the output of the intermediate layer as the \emph{intermediate feature}. The second part, which contains the remaining layers of the network, forms the classifier module. 
This way, using the layer separation mechanism, we achieve two essential objectives simultaneously: (i) obtaining the desired feature extractor, and (ii) benefiting from the intrinsic characteristics of DNNs: The output of higher layers are more specific to the main task, containing less irrelevant information.

If we just rely on the layer separation mechanism, we will end up with an embedding method, which we call \emph{Simple Embedding}, as illustrated in Fig~\ref{fig:simple}. In this embedding method, the feature extractor module that is sent to the user's device is just the initial layers of the network up to the intermediate layer with no changes, with the exclusive feature being the same as the intermediate feature. 

Choosing the intermediate layer from lower layers of the network intrinsically results in privacy compromises. As we proceed through the deep network layers, the feature becomes more specific to the main task, and irrelevant information (including sensitive information) will be gradually lost \cite{yosinski2014}. Meanwhile, the more privacy we gain by choosing higher layers as the intermediate one, the more processing overhead we impose on the user's device. Therefore, for the sake of scalability, it is better to select the intermediate layer from the lower layers of the network. However, lower layers of a DNN learn general invariant features that are not specific to the main task \cite{bengio2012deep}. This will bring privacy concerns with the simple embedding approach. The solution is to manipulate the intermediate feature coming from a lower intermediate layer and try to adjust it only for the desired primary measure so that all the other sensitive measures become unpredictable. To this end, we employ three different techniques: dimensionality reduction, noise addition, and Siamese fine-tuning, which are discussed accordingly.

\subsection{Dimensionality Reduction}
The first approach to increase the privacy of the intermediate feature is to reduce its dimensionality using Principle Component Analysis (PCA) or autoencoder-based solutions \cite{malekzadeh2017}, as these methods try to preserve the primary structure of the input signal as much as possible, and remove all the other unnecessary details. This mechanism also reduces the communication overhead between the user and the cloud. We only intended to show the effect of reducing dimensionality on the privacy of the exclusive features, so we simply used the popular PCA as a proof of concept. In this approach, a dense reduction matrix is added as the last layer of the feature extractor module, and a dense reconstruction matrix is attached before the first layer of the classification module. We refer to this embedding (with PCA applied) as \emph{reduced embedding}. In this method, the exclusive feature is obtained by reducing the dimensionality of the intermediate feature. As will be shown in Section~\ref{experiments}, this procedure does not significantly affect the accuracy of the main task. 



\subsection{Siamese Fine-tunning}

A more complicated approach to craft the exclusive feature is to utilize a many to one mapping for any sensitive variable. This is similar to the main idea behind \emph{k-anonymity}~\cite{sweeney2002}, when the identities are assumed to be sensitive variables. As an example, consider the problem of gender classification using a feature extractor that maps input images to a feature space. If we have $k$ images of the ``male'' class mapped to $k$ distinct points, an adversary may be able to reconstruct the original images if he could discover the reverse mapping function. Conversely, if all of the $k$ different male images map to a single point in the feature space, the adversary will experience confusion to select the correct identity between $k$ possible ones. Therefore, if we fine-tune the feature extractor in such a way that the features of the same class fall within a very small neighborhood of each other, the privacy of the input data will be better preserved. To accomplish this task, we use the Siamese architecture~\cite{chopra2005} to fine-tune the pre-trained model on the cloud, before applying the layer separation. 

\begin{figure}[t]
	\centering
	\begin{subfigure}[b]{0.48\columnwidth}
		\centering
		\includegraphics[width=.9\columnwidth]{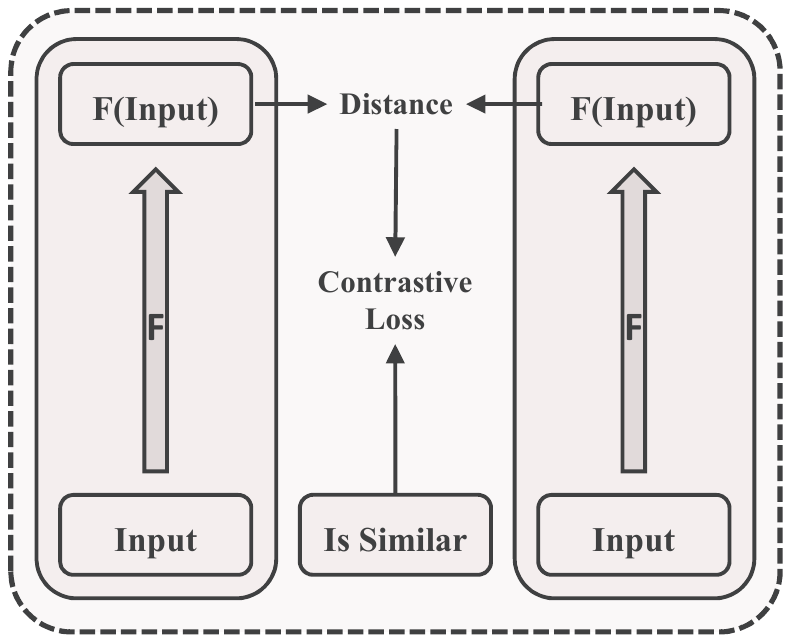}
		\captionsetup{format=hang}
		\caption{Traditional Siamese}
		\label{fig:siamese}
	\end{subfigure}%
	\quad
	\begin{subfigure}[b]{0.48\columnwidth}
		\centering
		\includegraphics[width=.9\columnwidth]{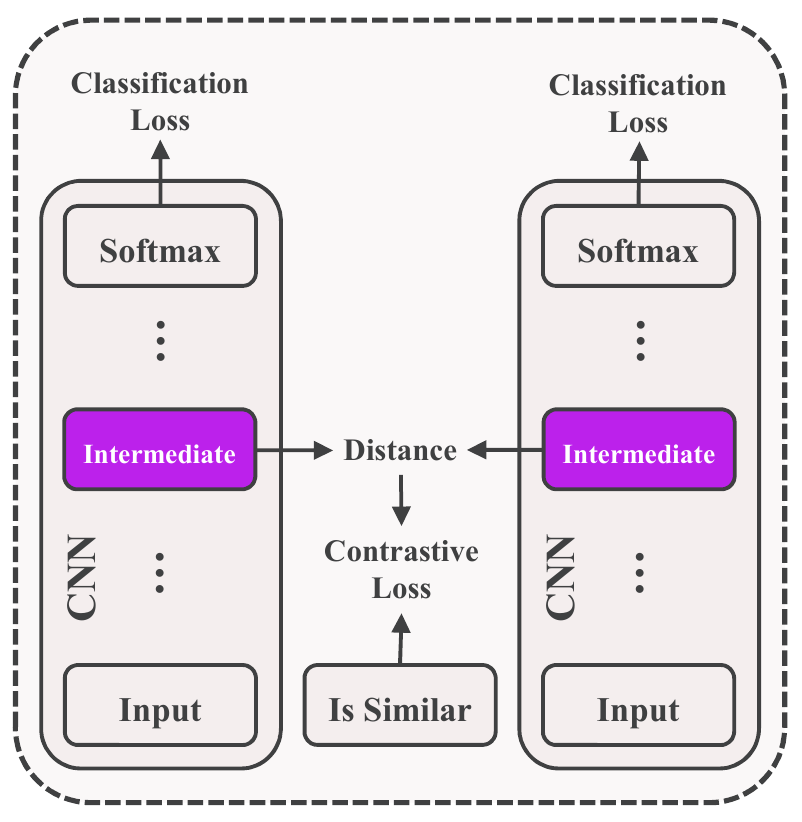}
		\captionsetup{format=hang}
		\caption{Privacy-preserving Siamese}
		\label{fig:siam_priv}
	\end{subfigure}%
	\caption{The figure on the left shows the traditional Siamese architecture used for some applications like face verification. The one on the right shows the privacy-preserving Siamese architecture used to fine-tune a pre-trained DNN on the cloud.}
	\label{fig:siamese_model}
\end{figure}

The Siamese architecture presented in Fig~\ref{fig:siamese_model} has previously been used in verification applications~\cite{chopra2005}. It provides a feature space where the similarity between data points is defined by their Euclidean distance. The main idea of fine-tuning with Siamese architecture is making the representation of semantically similar points become as close as possible to each other, while the representation of dissimilar points fall far from each other. 
For example, in face verification problem, where the goal is to determine whether two images belong to the same person or not, the Siamese fine-tuning can make any two images that belong to the same person fall within a local neighborhood in the feature space, and the features of any two images having mismatching faces become far from each other.

To fine-tune with Siamese architecture, the training dataset must comprise pairs of samples labeled as similar/dissimilar. A contrastive loss function is applied to each pair, such that the distance between two points gets minimized if they are similar and maximized otherwise. Here, we use the one introduced in~\cite{hadsell2006}:
\begin{align}
L(f_1,f_2) = \begin{cases}
\norm{f_1-f_2}_2^2 & \quad \text{\small similar}\\
\max(0,margin - \norm{f_1-f_2}_2)^2 & \quad \text{\small dissimilar}
\end{cases}
\end{align}
where $f_1$ and $f_2$ are the mappings of data points, and $margin$ is a hyper parameter controlling the variance of the feature space.

\begin{figure}[t]
	\centering
	\begin{subfigure}[b]{.48\columnwidth}
		\centering
		\includegraphics[width=.9\columnwidth]{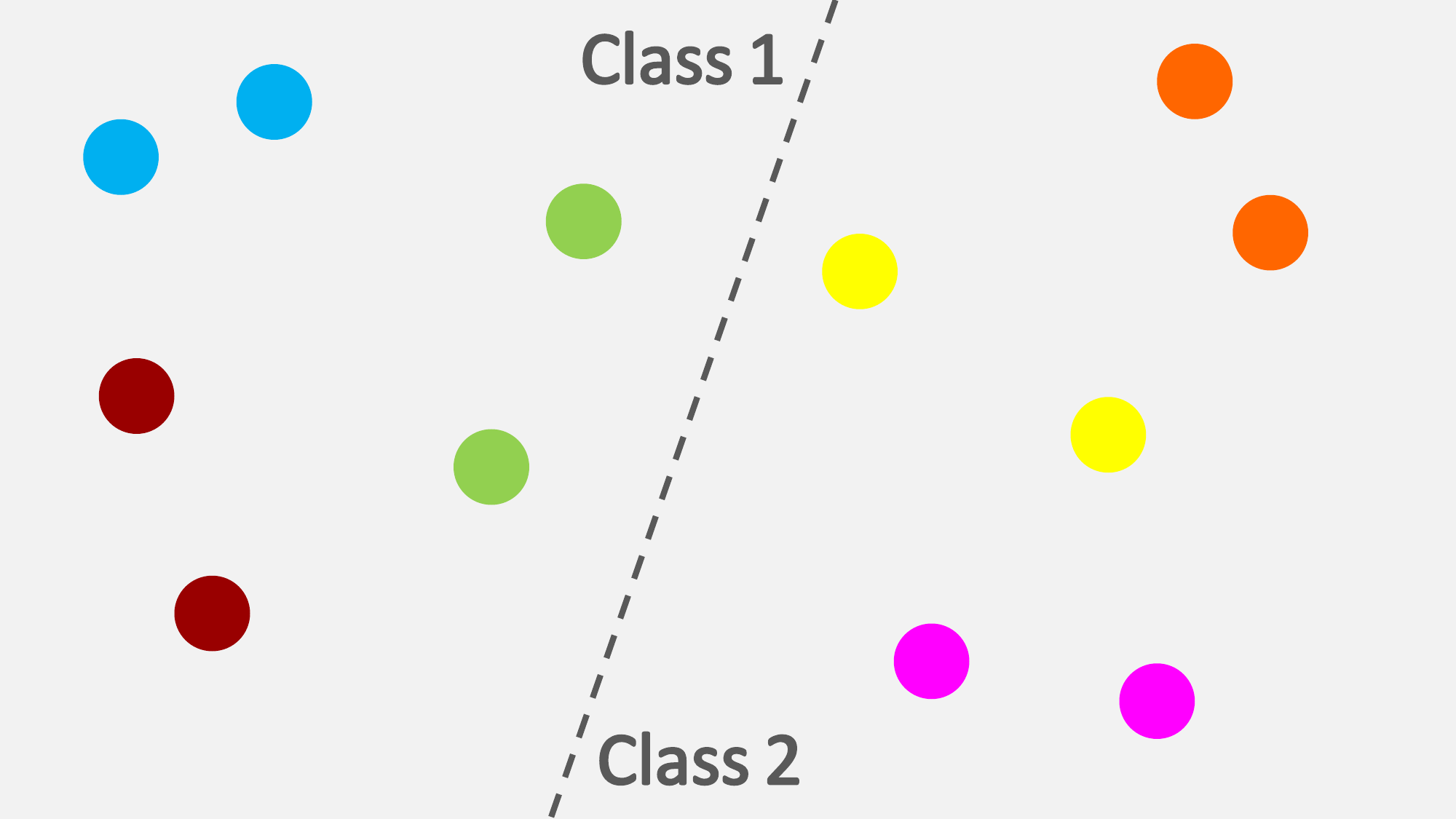}
		\captionsetup{format=hang}
		\caption{Before fine-tuning}
		\label{fig:before-siamese}
	\end{subfigure}%
	\quad
	\begin{subfigure}[b]{.48\columnwidth}
		\centering
		\includegraphics[width=.9\columnwidth]{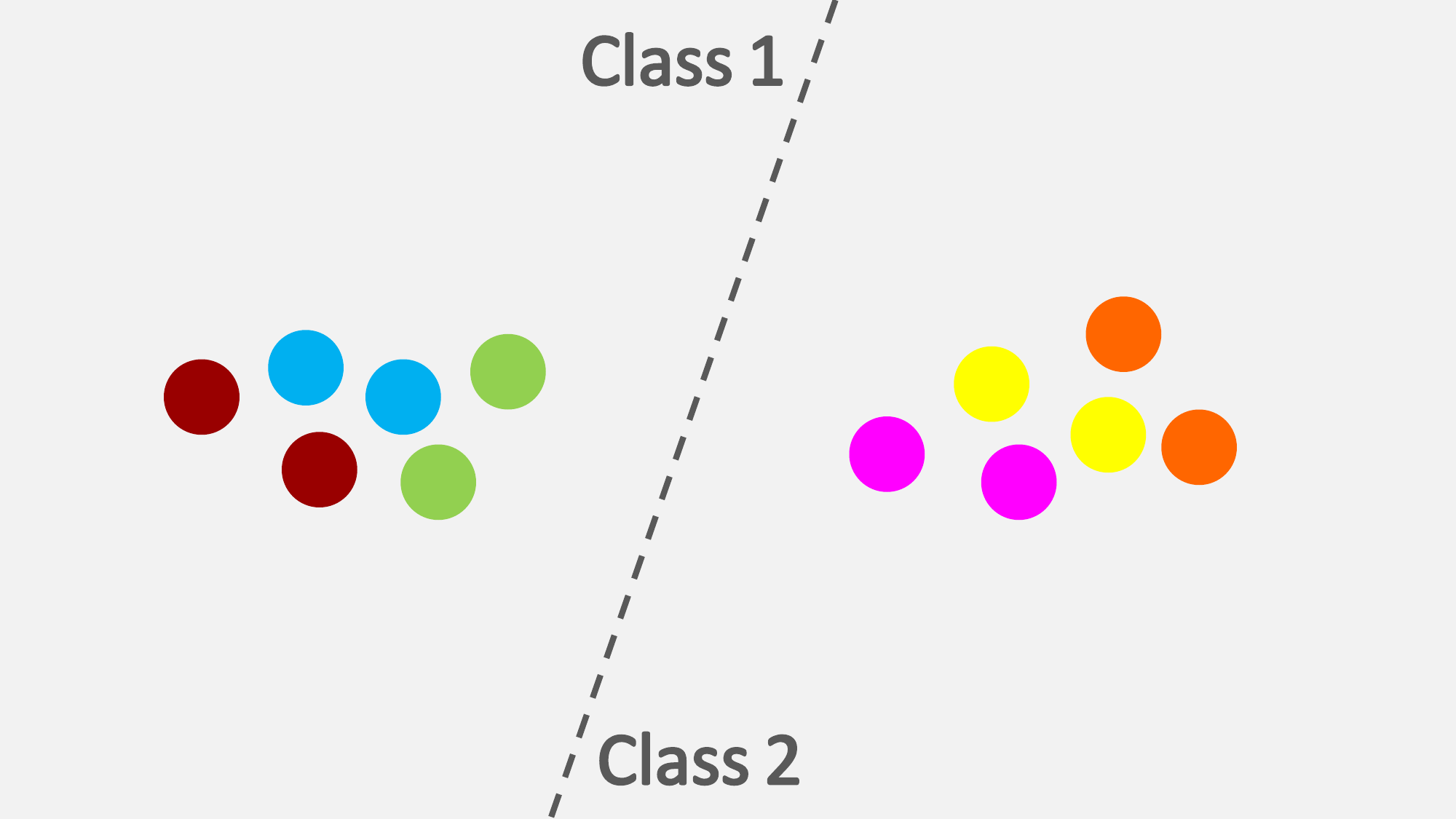}
		\captionsetup{format=hang}
		\caption{After fine-tuning}
		\label{fig:after-siamese}
	\end{subfigure}%
	\caption{{A two-dimensional representation of the feature space before and after Siamese fine-tuning where the main task is a two-class classification. The dashed line shows the border between primary class 1 and 2. The points with the same color have the same attribute, which is potentially sensitive. Before Siamese fine-tuning, an adversary can discover sensitive information as there are clear borders between different colors, but after fine-tuning, the sample points with the same primary class will get too close to each other, making the inference of the sensitive information much harder. A small perturbation of the data points after the fine-tuning will totally dispose of the sensitive class borders while maintaining them on the right-hand side of the primary class border.}}
	\label{fig:siamese-effect}
\end{figure}

{We exploit Siamese fine-tuning to increase the privacy of the feature extractor by forcing it to map the input data to the feature space in such a way that samples with the same primary class label become close to each other, while samples with different class labels become far. This is illustrated in Fig.~\ref{fig:siamese-effect}. For example, if the main task is gender recognition, with Siamese fine-tuning, all the images of the "male" class will be mapped to a small local region in the feature space, and similarly, all the "female" images will be mapped to another region which is far from the male's.}

{By defining a contrastive loss on the intermediate layer, we get a multi-objective optimization problem. It tries to increase the accuracy of the primary variable prediction by minimizing the classification loss, and at the same time, it increases the privacy of the exclusive feature by minimizing the contrastive loss. In fact, we add the contrastive loss as a regularization term to the main classification loss (weighted sum) and try to optimize the new loss function with gradient descent based algorithms.}


We refer to this embedding method as \emph{Siamese embedding}. Note that the whole process of Siamese fine-tuning is done in the cloud by the service provider only once, before applying the layer separation and delivering the feature extractor module to end-users. The exclusive feature, in this case, will be the output of the intermediate layer of the fine-tuned DNN.

\subsection{Noise Addition}

%

Noise addition is another traditional method for preserving privacy \cite{agrawal2000}, which increases the inference uncertainty of unauthorized tasks. 
Apart from Siamese fine-tuning and dimensionality reduction, the feature extractor module can also add multi-dimensional noise to the feature vector to further increase the privacy. We refer to this technique as \emph{noisy embedding}. {Although Siamese fine-tuning tries to map data points with different sensitive classes to a single point, these points may still have small distances from each other (Fig~\ref{fig:after-siamese}). We can highly increase the uncertainty about sensitive variables by adding random noise to the feature.}
As the variance of the added noise increases, we get more uncertainty in the sensitive variable, and thus the privacy will be better preserved. However, a high-variance noise could also decrease the prediction accuracy of the primary variable, as it could cause the data points to fall out of the right-class region.
Therefore, we have a trade-off between privacy and accuracy when increasing the amount of noise. A significant benefit of Siamese fine-tuning is that it allows us to add noise with higher variance, because after the fine-tuning, the intra-class variance of the primary variable decreases, while the inter-class variance gets increased. Therefore, with Siamese fine-tuning, our framework can tolerate higher variance noises without noticeable performance drop in the main task. 

\begin{figure}[t]
	\centering
	\includegraphics[width=.95\columnwidth]{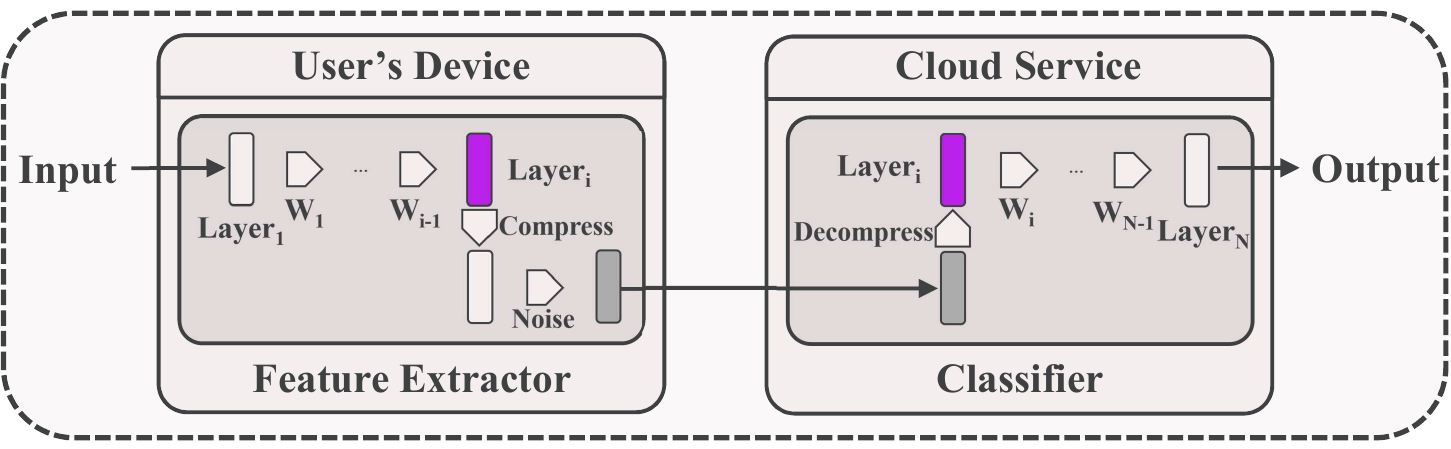}
	\caption{The noisy reduced embedding of a pre-trained DNN with both dimensionality reduction and noise addition. If Siamese fine-tuning is also applied to the DNN before layer separation in the cloud, it will be the advanced embedding method. In this case, after cloud-side Siamese fine-tuning, the feature extractor is sent to the user's device where PCA projection and noise addition are applied to produce the exclusive feature. This feature is then uploaded to the cloud to proceed with the processing.}
	\label{fig:advanced}
\end{figure}

Ultimately, we can combine all the embedding methods, namely Siamese fine-tuning with both dimensionality reduction and noise addition, as a method that we call \textit{advanced embedding}. Fig~\ref{fig:advanced} illustrates an overview of the advanced embedding, in which Siamese fine-tuning is performed on the cloud before pushing the feature extractor to the end-users' devices. When the user attempts to use the service, the feature extractor module on her device extracts a feature vector from the input data at the intermediate layer. Next, the dimensionality of the obtained feature is reduced by applying a PCA or an Auto-Encoder. Before uploading to the cloud, the feature extractor adds some noise to the reduced-size feature to obtain the exclusive feature. The classifier module residing in the cloud receives the exclusive feature and performs a decompression operation. Finally, the reconstructed feature is fed into the neural network to produce the expected result.


%% file: evaluation.tex

\section{Privacy Verification}
\label{evaluation}

In this section, we introduce three different methods to verify and evaluate the privacy of the exclusive feature. First, we define a privacy evaluation measure based on a statistical analysis of an arbitrary sensitive variable given the exclusive feature. Next, we use transfer learning~\cite{yosinski2014} to determine the degree of generality and particularity of the exclusive feature to the main task, and deep visualization~\cite{dosovitskiy2016} to evaluate how well we can reconstruct input data from the exclusive feature. In practice, these methods can be used by the end-user or a third party to verify the privacy of the feature extractor module of a service provider in order to decide if the service provider should be trusted.


\subsection{Privacy Measure}
\label{validation}
Here, we introduce an intuitive method to measure and quantify the privacy of the feature extractor. We consider the cases where we want to verify the privacy of the exclusive feature concerning a discrete sensitive variable, such as the identity of people. Although there exist various methodologies for this purpose in the literature, including k-anonymity~\cite{sweeney2002} or differential privacy~\cite{dwork06}, they can not be directly applied to our problem, since they are designed for different privacy-preserving scenarios such as dataset and model publishing (discussed in Section~\ref{sec:related}). However, as our method addresses the privacy issue in a different context, we have to evaluate the privacy of our framework using a different measure. 

The privacy of our framework directly relies on the amount of sensitive information that exists in the exclusive feature. The reliable approach to this end is to use information-theoretic concepts such as mutual information or conditional entropy to measure the amount of sensitive information. However, even if we were able to obtain a reasonable estimate of the joint distribution of the sensitive variable and the exclusive feature, calculating these measures would be intractable, especially in the high dimensional space \cite{haykin2009}. Another approach is to consider the deficiency of the classifier trying to discriminate an arbitrary sensitive variable, given the exclusive feature. In this context, less classification accuracy implies that the exclusive feature contains less information about the sensitive variable, and thus more privacy is preserved. However, classification accuracy by itself is not sufficient to guarantee privacy because the actual sensitive class might be among the most probable candidates. Therefore, we extend this approach to include the rank of the results in the measure. Considering the rank of likelihood as the privacy measure makes sense, because it can be seen as an extension of k-anonymity and top-k accuracy, and also an estimation of guessing entropy \cite{massey1994guessing}. In fact, average rank is equivalent to the expected k-anonymity, and lower ranks result in less top-k accuracy and thus more privacy. Moreover, the average rank is an empirical estimation of guessing entropy, which is a well-known measure of uncertainty \cite{massey1994guessing} that can also be used as a measure of privacy.

Formally speaking, assume that we have a dataset $D=\{(x_i,s_i)\}_{i=1:N}$, where $x_i$ is an input data and $s_i$ is a discrete sensitive class which can take values from the set $\{1,2,\dots,K\}$. We apply the feature extractor on $D$ to obtain the set of features $F=\{(f_i,s_i)\}_{i=1:N}$. Then, by adding noise to $F$, we build the set of noisy features $Z=\{(z_i,s_i)\}_{i=1:N}$. To measure the privacy of a certain noisy feature $z_i$, we calculate the conditional likelihood of all the sensitive classes $\{P(s|z_i)\ |\ 1\leq s\leq K\}$. Therefore, we first estimate $P(z_i | s)$ for an arbitrary sensitive class $s$ as:

\begin{equation}
P(z_i | s)  = \int_f P(z_i,f | s) df  = \int_f P(z_i | f,s)P(f | s) df
\end{equation}
if we condition on $f$, $s$ becomes independent of $z_i$, and we have:
\begin{equation}
P(z_i | s)  = \int_f P(z_i | f)P(f | s) df = E_{f \sim P(f|s)} [P(z_i | f)]
\end{equation}
Assuming $F_s=\{f_1, f_2, ... , f_{N_s}\}$ is the set of extracted features with the sensitive class $s$ in our dataset, we can estimate the above expected value with sample mean calculated on $F_s$:
\begin{equation}
\hat{P}(z_i|s) = \frac{1}{N_{s}} \sum_{f_j \in F_s}P(z_i | f_j)
\end{equation}
Now by using the Bayes rule, we have:
\begin{equation}
\hat{P}(s|z_i) \propto \hat{P}(z_i|s)P(s)
\end{equation}

Then, we can compute the relative likelihood of all sensitive classes given a noisy feature $z_i$. As we know the correct sensitive class of $z_i$ which is $s_i$, we find the \emph{rank} of the likelihood of the true class $P(s_i|z_i)$ in the set $\{P(s|z_i) | 1\leq s\leq N_s\}$ sorted in descending order (lower likelihoods get higher rank), as the privacy of $z_i$:
\begin{equation}
Privacy(z_i)=\frac{Rank(s_i)}{K}
\end{equation}
where we divide the rank by $K$ (the total number of classes) to normalize the values between 0 and 1.
For all of the members of $Z$, we can estimate the total privacy of the transmitted data by averaging over the individuals privacy values:
\begin{equation}
Privacy_{\ total} = \frac{1}{N}\sum_{i=1}^N Privacy(z_i)
\end{equation}

{The utility-privacy trade-off for this measure is addressed in \cite{osia2019privacy} from information-theoretic point of view. In this paper we use this measure in practice and create the accuracy-privacy trade-off (see Fig \ref{fig:gc_accpriv}).}


\subsection{Transfer Learning}
\label{transfer}

\begin{figure}[t]
	\centering
	\includegraphics[width=.7\columnwidth]{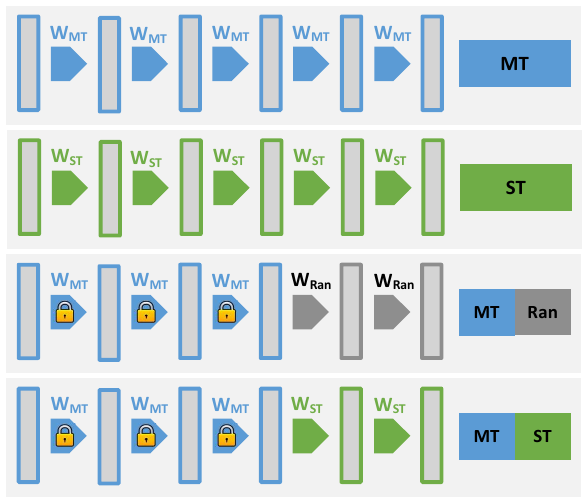}
	\caption{The transfer learning procedure from top to bottom: (1) the trained network $N_1$ for the main task; (2) another network $N_2$ for the secondary task; (3) the weights of the beginning layers are copied from $N_1$ to $N_2$ and frozen, while the remaining layers initialize with random weights; (4) final trained network for the secondary task, using transfer learning.}
	\label{fig:transfer}
\end{figure}

We can also measure the amount of particularity of the extracted feature to the main task using Transfer Learning~\cite{yosinski2014}. This procedure is depicted in Fig~\ref{fig:transfer}. Suppose we have already trained a DNN $N_1$ for primary variable classification. Then, we build and train another DNN $N_2$ to infer an arbitrary sensitive variable according to the following procedure:
\begin{enumerate}
	\item Copy weights from the first $i$ layers of $N_1$ to the first $i$ layers of $N_2$.
	\item Initialize the remaining layers of $N_2$ randomly.
	\item Freeze the first $i$ layers of $N_2$ (do not update their weights).
	\item Train $N_2$ for sensitive variable inference to learn the rest of the parameters.
\end{enumerate}

After the training procedure, the accuracy obtained for sensitive variable prediction is directly related to the degree of generality of the extracted feature from $i$'th layer. The lower accuracy for the sensitive variable prediction means the more specific is the feature to the main task.

\subsection{Deep Visualization}

Visualization is a method for understanding deep networks. Deep visualization can give us a fascinating insight into the privacy of the exclusive feature for those tasks in which the input is visualizable, such as images. In order to obtain an insight into the amount of sensitive information in the exclusive feature, we use an Auto-Encoder visualization technique \cite{dosovitskiy2016}, which is especially useful when working with image datasets. In \cite{dosovitskiy2016}, a decoder is designed on the data representation of each layer to reconstruct the original input image from the learned representation. Therefore, we can analyze the preserved sensitive information in each layer by comparing the reconstructed image with the original input.

%% file: experiments.tex
\section{Experiments}
\label{experiments}

In this section, we evaluate and analyze the accuracy and privacy of different embedding methods for two widely-used classification tasks as a case study: activity recognition and gender classification. For activity recognition, we consider gender as a piece of arbitrary sensitive information that the user does not want to be exposed to the cloud service provider, whereas for gender classification, we set the sensitive variable to identity. In each of these applications, we verify the feature extractor module by using the proposed privacy measure and transfer learning. We show that among all the proposed embedding methods, applying Siamese fine-tuning is the most efficient one in preserving privacy, yet it does not decrease the accuracy of the main classification task. Besides, we show how deep visualization is not feasible in reconstructing the original data from the exclusive feature. Finally, we evaluate our hybrid framework on a mobile phone and discuss its advantages compared to other solutions.

\subsection{Experiment Settings}

We embed the proposed framework into the state of the art models for gender classification and activity recognition and evaluate its effectiveness in preserving users' data privacy. 


\begin{figure}[t]
	\begin{center}
		\includegraphics[width=.8\columnwidth, height=4cm]{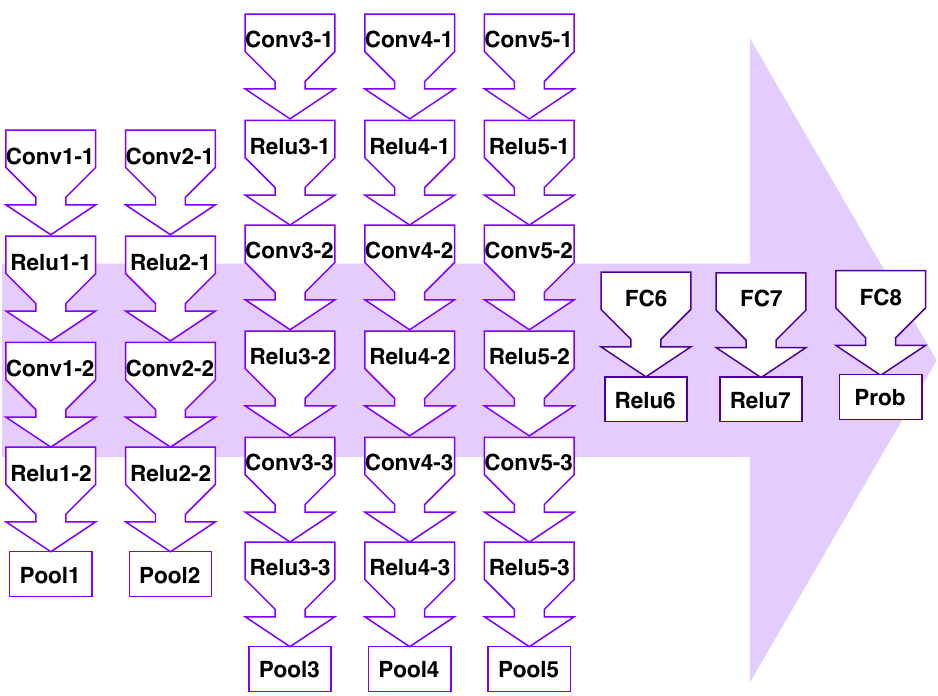} 	
	\end{center}
	\caption{VGG-16 architecture for gender classification \cite{simonyan2014very}.}
	\label{fig:vgg-16}
\end{figure}

\subsubsection{Gender Classification} In the problem of gender classification, the goal is to classify an individual's image to \emph{male} or \emph{female}, without disclosing the identity of individuals as an example of sensitive information. This has various applications in different domains such as human-computer interaction, surveillance, and targeted advertising systems~\cite{ng2012}. 

\descr{Datasets.} Rothe~\etal~\cite{rothe2015} prepared a large dataset, named IMDB-Wiki, which is useful for age and gender estimation. We used the Wiki part of this dataset, containing 62,359 images, to fine-tune the gender classification model with Siamese architecture. We used 45,000 images as training data and the rest as validation data. We evaluated our privacy measurement techniques on this dataset, as well. We also used ``Labeled Face in the Wild'' (LFW) dataset~\cite{LFWTech} as a benchmark to test the predictive accuracy of different embedding methods. This is an unconstrained face database containing 13,233 images of 5,749 individuals, which is very popular for evaluating face verification and gender classification models. 
For the transfer learning approach, we used the IMDB dataset from~\cite{parkhi2015}, containing about 2 million images from 2,622 well-known celebrities on the IMDB website, among which we selected 100 and divided their images to training and test sets for evaluating the face recognition model.

\descr{Setup.} We used the pre-trained model proposed in~\cite{rothe2015} having 94\% accuracy based on VGG-16 architecture~\cite{simonyan2014very}, which is shown in Fig~\ref{fig:vgg-16}. We broke it down on Conv5-1, Conv5-2, and Conv5-3 as different intermediate layers and evaluated our framework on each of them. To create reduced embeddings, we applied PCA on the intermediate feature to reduce its dimension to 4, 6, and 8 for Conv5-3, Conv5-2, and Conv5-1, respectively. We did not try to optimize the dimension setting aggressively and just used cross-validation to select some reasonable dimensions.

\begin{figure}[t]
	\begin{center}
		\includegraphics[width=.6\columnwidth]{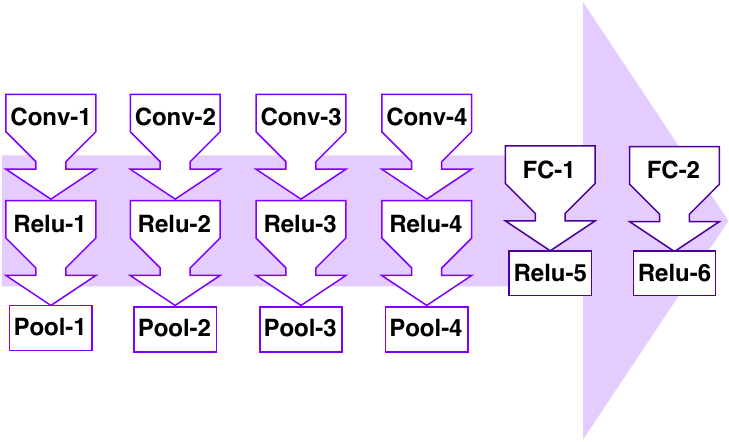} 	
	\end{center}
	\caption{MTCNN architecture for activity recognition \cite{malekzadeh2018protecting}.}
	\label{fig:mtcnn}
\end{figure}

\subsubsection{Activity Recognition} Another application we used for evaluating the proposed framework is activity recognition, which aims to recognize the activity of a user from the accelerometer and gyroscope data of her smartphone. In this scenario, we set to prevent the exposure of the user's gender as a piece of arbitrary sensitive information. 

\descr{Dataset.} We used the MotionSense dataset \cite{malekzadeh2018protecting} for all the stages of Siamese fine-tuning, predictive accuracy assessment, and transfer learning. This dataset contains time-series data generated by accelerometer and gyroscope sensors (attitude, gravity, user acceleration, and rotation rate), collected by an iPhone 6s kept in the 24 participant's front pocket in 15 trials doing the following activities: downstairs, upstairs, walking, and jogging.

\descr{Setup.} The activity recognition model we used in experiments is the already trained version of a specific implementation of a multi-task convolutional neural network (MTCNN) proposed by Malekzadeh \etal \cite{malekzadeh2018protecting}. The overall view of this architecture is depicted in Fig~\ref{fig:mtcnn}. We decoupled the network on Conv-4, FC-1, and FC-2 as different intermediate layers. Reduced embeddings of this model were obtained from applying PCA to reduce the dimension of the intermediate feature to 8 for each of the layers.

\subsection{Experiment Results}

\begin{table}[t]
	\caption{Predictive Accuracy of Different Embedding Methods}
\scriptsize
		\begin{tabu} to \columnwidth { X[l] X[c] X[c] X[c] c X[c] X[c] X[c]}
			\toprule
			& \multicolumn{3}{c}{Gender Classification} & & \multicolumn{3}{c}{Activity Recognition} \\
			\cmidrule{2-4}  \cmidrule{6-8}
			Method & Conv5\nobreakdash-1 & Conv5\nobreakdash-2 & Conv5\nobreakdash-3 & & Conv\nobreakdash-4 & FC-1 & FC-2 \\
			\midrule

			Simple  & 94.0\% & 94.0\%  &  94.0\% && 93.0\% & 92.7\%& 93.2\% \\ 
			Reduced & 89.7\%  &  87.0\% & 94.0\% && 85.3\%& 92.5\%& 93.1\% \\
			Siamese & 92.7\% & 92.7\%  &  93.5\% && 93.2\%& 93.3\%& 94.2\% \\
			Red.~Siam.& 91.3\% &  92.9\% &  93.3\% && 90.1\%& 92.8\%& 94.2\% \\
			\bottomrule
		\end{tabu}
	\label{tab:gcpca}
\end{table}

\descr{Accuracy of Embedding Methods.} In the first step, we assessed how different embedding methods affect the accuracy of both gender classification and activity recognition as two different main tasks. Table~\ref{tab:gcpca} reports the accuracy of both tasks under different embedding methods and intermediate layers {(the accuracy of noisy version can be seen in Fig~\ref{fig:gc_accpriv} for different amounts of noise)}. The obtained results convey two critical messages. First, the predictive accuracy of Siamese and simple embeddings are comparable, which implies that applying Siamese fine-tuning does not necessarily decrease the accuracy of the main task, but sometimes can even result in a slight improvement due to the way it centralizes same-class data points and separates different-class ones. Second, Siamese embedding is more robust to PCA than simple embedding, again due to the same reason. In other words, the accuracy of Siamese embedding is very close to its reduced counterpart, while for simple embedding, applying dimensionality reduction will deteriorate the accuracy. As a result, applying the Siamese embedding with dimensionality reduction leads to better privacy with less accuracy degradation of the main task.

%

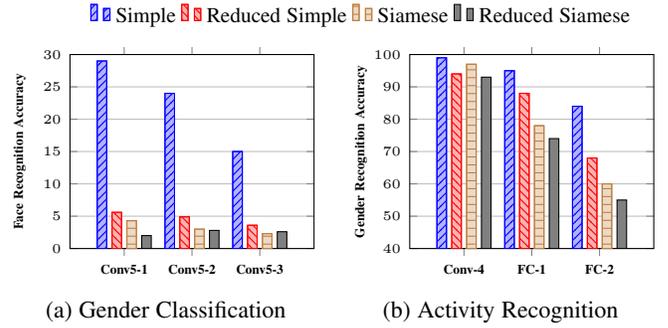
\begin{figure}[t]
	\definecolor{color1}{HTML}{B2B2FF}
	\definecolor{color2}{HTML}{FFB2B2}
	\definecolor{color3}{HTML}{ECD9C6}
	\definecolor{color4}{HTML}{808080}
	\centering
	\begin{subfigure}[t]{.48\columnwidth}
		\centering
		\begin{tikzpicture}
		\begin{axis}[
		ybar,
		tiny,
		width=1.9in,
		enlarge x limits=0.4,
		bar width=0.05in,
		legend style={at={(1.2,1.3)},font=\footnotesize,
			anchor=north,legend columns=-1, draw=none},
		symbolic x coords={Conv5-1, Conv5-2, Conv5-3},
		xtick=data,
		xticklabels={\textbf{Conv5-1}, \textbf{Conv5-2}, \textbf{Conv5-3}},
		ymin=0,ymax=30,
		ylabel=\textbf{Face Recognition Accuracy},
		ymajorgrids,
		y tick label style={
			/pgf/number format/.cd,
			fixed,
			fixed zerofill,
			precision=0,
			/tikz/.cd
		},
		]
		\addplot[fill=color1,draw=blue,postaction={pattern=north east lines,pattern color=blue}] coordinates {(Conv5-1,29) (Conv5-2,24) (Conv5-3,15)};
		\addplot[fill=color2,draw=red,postaction={pattern=north west lines,pattern color=red}] coordinates {(Conv5-1,5.6) (Conv5-2,4.9) (Conv5-3,3.6)};
		\addplot [fill=color3,draw=brown,postaction={pattern=horizontal lines,pattern color=brown}] coordinates {(Conv5-1,4.3) (Conv5-2,3) (Conv5-3,2.3)};
		\addplot [fill=color4,draw=black] coordinates {(Conv5-1,2.0) (Conv5-2,2.8) (Conv5-3,2.6)};
		
		\legend{Simple,Reduced Simple, Siamese, Reduced Siamese}
		\end{axis}
		\end{tikzpicture}
		\caption{Gender Classification}
		\label{fig:FRGC}
		\vspace*{15pt}
	\end{subfigure}
\hfil
	\begin{subfigure}[t]{.48\columnwidth}
		\centering
		\begin{tikzpicture}
		\begin{axis}[
		ybar,
		tiny,
		width=1.9in,
		enlarge x limits=0.4,
		bar width=0.05in,
		symbolic x coords={Conv-4, FC-1, FC-2},
		xtick=data,
		xticklabels={\textbf{Conv-4}, \textbf{FC-1}, \textbf{FC-2}},
		ymin=40,ymax=100,
		ylabel=\textbf{Gender Recognition Accuracy},
		ylabel shift=-4pt,
		ymajorgrids,
		y tick label style={
			/pgf/number format/.cd,
			fixed,
			fixed zerofill,
			precision=0,
			/tikz/.cd
		},
		]
		\addplot [fill=color1,draw=blue,postaction={pattern=north east lines,pattern color=blue}] coordinates {(Conv-4,99) (FC-1,95) (FC-2,84)};
		\addplot [fill=color2,draw=red,postaction={pattern=north west lines,pattern color=red}] coordinates {(Conv-4,94) (FC-1,88) (FC-2,68)};
		\addplot [fill=color3,draw=brown,postaction={pattern=horizontal lines,pattern color=brown}] coordinates {(Conv-4,97) (FC-1,78) (FC-2,60)};
		\addplot [fill=color4,draw=black] coordinates {(Conv-4,93) (FC-1,74) (FC-2,55)};
		
		\end{axis}
		\end{tikzpicture}
		\caption{Activity Recognition}
		\label{fig:FRAC}
	\end{subfigure}

	\caption{Transfer learning results for different embeddings across different intermediate layers. Less accuracy means more privacy. {According to the figure, using higher layers together with Siamese fine-tuning and dimensionality reduction achieves more privacy.}}
	\label{fig:trans}
\end{figure}

\begin{figure*}[t]
	\centering
	\begin{subfigure}{0.2\textwidth}
		\begin{tikzpicture}
		\begin{axis}
		[
		tiny,
		width=4.5cm,
		height=4.5cm,
		legend pos=north east,
		legend style={nodes={scale=0.75, transform shape}, legend columns=2},
		legend image post style={scale=0.5},
		grid,
		y tick label style={
			/pgf/number format/.cd,
			fixed,
			fixed zerofill,
			precision=0,
			/tikz/.cd
		},
		xlabel=\textbf{Identity Privacy (\%)},
		ylabel=\textbf{Gender Classification Accuracy (\%)},
		ylabel shift = -4 pt,
		ymax=96, ymin=84,
		xmin=0,
		xmax=40,
		legend entries={\textbf{Noisy Reduced}, \textbf{Advanced}},
		]
		\addplot[color=purple,mark=square*,mark size=1.1,thick] table{simple_gender_conv53_rank.txt}; 
		\addplot[color=cyan,mark=*,mark size=1.1,thick] table{siamese_gender_conv53_rank.txt};
		\end{axis}
		\end{tikzpicture}
		\caption{\small{Effect of Siamese fine-tunning on gender classification using conv5-3 as intermediate layer}}
		\label{fig:gender_presence}
	\end{subfigure}%
\hfil
	\begin{subfigure}{0.2\textwidth}
		\begin{tikzpicture}
		\begin{axis}
		[
		tiny,
		width=4.5cm,
		height=4.5cm,
		legend pos=north east,
		legend style={nodes={scale=0.75, transform shape},legend columns=3},
		legend image post style={scale=0.5},
		grid,
		y tick label style={
			/pgf/number format/.cd,
			fixed,
			fixed zerofill,
			precision=0,
			/tikz/.cd
		},
		xlabel=\textbf{Identity Privacy (\%)},
		ylabel=\textbf{Gender Classification Accuracy (\%)},
		ylabel shift = -4 pt,
		ymax=96, ymin=84,
		xmin=0,
		xmax=40,
		legend entries={\textbf{Conv5-1}, \textbf{Conv5-2}, \textbf{Conv5-3}},
		]
		\addplot[color=orange,mark=triangle*,mark size=1.5,thick] table{siamese_gender_conv51_rank.txt}; 
		\addplot[color=cyan,mark=*,mark size=1.1,thick] table{siamese_gender_conv52_rank.txt}; 
		\addplot[color=purple,mark=square*,mark size=1.1,thick] table{siamese_gender_conv53_rank.txt}; 
		\end{axis}
		\end{tikzpicture}
		\caption{\small{Effect of different intermediate layers on gender classification using advanced embedding}}
		\label{fig:gender_layers}
	\end{subfigure}%
\hfil
	\begin{subfigure}{.2\textwidth}
		\begin{tikzpicture}
		\begin{axis}
		[
		tiny,
		width=4.5cm,
		height=4.5cm,
		legend pos=north east,
		legend style={nodes={scale=0.75, transform shape}, legend columns=2},
		legend image post style={scale=0.5},
		grid,
		y tick label style={
			/pgf/number format/.cd,
			fixed,
			fixed zerofill,
			precision=0,
			/tikz/.cd
		},
		xlabel=\textbf{Gender Privacy (\%)},
		ylabel=\textbf{Activity Recognition Accuracy (\%)},
		ylabel shift = -4 pt,
		ymax=96, ymin=84,
		xmin=0,
		xmax=40,
		legend entries={\textbf{Noisy Reduced}, \textbf{Advanced}},
		]
		\addplot[color=purple,mark=square*,mark size=1.1,thick] table{simple_activity_1.txt}; 
		\addplot[color=cyan,mark=*,mark size=1.1,thick] table{siamese_activity_11.txt};
		\end{axis}
		\end{tikzpicture}
		\caption{Effect of Siamese fine-tunning on activity recognition using FC-2 as intermediate layer}
		\label{fig:activity_presence}	
	\end{subfigure}
\hfil
	\begin{subfigure}{.2\textwidth}
		\begin{tikzpicture}
		\begin{axis}
		[
		tiny,
		width=4.5cm,
		height=4.5cm,
		legend pos=north east,
		legend style={nodes={scale=0.75, transform shape}, legend columns=3},
		legend image post style={scale=0.5},
		grid,
		y tick label style={
			/pgf/number format/.cd,
			fixed,
			fixed zerofill,
			precision=0,
			/tikz/.cd
		},
		xlabel=\textbf{Gender Privacy (\%)},
		ylabel=\textbf{Activity Recognition Accuracy (\%)},
		ylabel shift = -4 pt,
		ymax=96, ymin=84,
		xmin=0,
		xmax=40,
		legend entries={\textbf{Conv-4}, \textbf{FC-1}, \textbf{FC-2}},
		]
				\addplot[color=orange,mark=triangle*,mark size=1.5,thick] table{siamese_activity_3.txt};
		\addplot[color=cyan,mark=*,mark size=1.1,thick] table{siamese_activity_2.txt}; 
		\addplot[color=purple,mark=square*,mark size=1.1,thick] table{siamese_activity_11.txt};
		\end{axis}
		\end{tikzpicture}
		\caption{\small{Effect of different intermediate layers on activity recognition using advanced embedding}}
		\label{fig:activity_layers}	
	\end{subfigure}

	\caption{{Accuracy-privacy trade-off. In each plot, y-axis shows accuracy of the main task (gender classification in (a) and (b), and activity recognition in (c) and (d)), and x-axis shows privacy of the sensitive task (identity recognition in (a) and (b), and gender recognition in (c) and (d)). For the left-most point in each plot, noise is zero, and it gains the most accuracy with the least privacy. By increasing the amount of noise variance, we can obtain more privacy while giving up some accuracy. Each point represents a certain amount of variance for the added noise. We can observe that the advanced embedding (which includes Siamese fine-tuning) achieves better trade-off than the noisy reduced embedding (without Siamese fine-tuning) (see (a) and (c)). Also, higher intermediate layers lead to better trade-off (see (b) and (d)).}}
	\label{fig:gc_accpriv}

\end{figure*}

\descr{Transfer Learning.}
The result of transfer learning for different embeddings on different intermediate layers for both gender classification and activity recognition tasks are presented in Fig~\ref{fig:trans}. For gender classification (Fig~\ref{fig:FRGC}), applying (reduced) simple or Siamese embedding results in a considerable decrease in the accuracy of face recognition (secondary task) from Conv5-1 to Conv5-3. The reason for this trend is that as we go up through the layers, the features of each layer will be more specific to the main task. In other words, the feature of each layer will have less information related to identity (the sensitive information) comparing to the previous ones. In addition, the face recognition accuracy of Siamese embedding is by far less than the accuracy of simple embedding in all configurations. As it is shown in Fig~\ref{fig:FRGC}, when Conv5-3 is chosen as the intermediate layer in Siamese embedding, the accuracy of face recognition is 2.3\%. Another interesting point of Fig~\ref{fig:trans} is the effect of dimensionality reduction on the accuracy of face recognition. Reduced simple and reduced Siamese embeddings have lower face recognition accuracy than simple and Siamese embeddings, respectively. 

We observe similar trends for the activity recognition task depicted in Fig~\ref{fig:FRAC}, in which the accuracy of the gender recognition model (secondary task) is decreased in all embedding methods across different intermediate layers. Analogous to gender classification, the Siamese embedding works better than simple embedding for activity recognition as well, and applying dimensionality reduction helps with better privacy protection.

\descr{Privacy Measure.}
We used the rank measure proposed in Section~\ref{validation} to evaluate the privacy of different embedding methods against their achieved accuracy for the gender classification and activity recognition tasks. We increased the privacy of noisy embedding methods by widening the variance of the added symmetric Gaussian noise. Fig~\ref{fig:gc_accpriv} illustrates the achieved privacy based on this measure against the predictive accuracy of the main task for different noisy configurations of gender classification and activity recognition. Each data point in the figures represents a certain amount of variance for the added noise.

In Fig~\ref{fig:gender_presence}, for the gender classification task, we fixed the intermediate layer on Conv5-3 and evaluated the accuracy-privacy trend of two embedding methods: noisy reduced simple and advanced embeddings in order to evaluate the effect of Siamese fine-tuning. We can see from the figure that as the privacy increases (by adding higher-variance noises), gender classification accuracy decreases more slowly in advanced embedding, where we use Siamese fine-tuning, compared to the other method that does not benefit from Siamese fine-tuning. 
We also assessed the effect of choosing different intermediate layers on the accuracy-privacy curve when using the advanced embedding method. The result of this experiment is depicted in Fig~\ref{fig:gender_layers}, where the accuracy-privacy curve of higher layers are above lower ones, meaning that higher layers provide more privacy when achieving the same accuracy. This result conforms with that of transfer learning in a way that choosing the intermediate layer closer to the output of the network results in having a lower face recognition accuracy.

Similar to gender classification, we also evaluated the accuracy-privacy trade-off for different embedding techniques in the activity recognition task. The effect of having Siamese fine-tuning is shown in Fig~\ref{fig:activity_presence}, where we fixed the intermediate layer to FC-2. As expected, the advanced embedding curve is above the one for noisy reduced simple embedding in this task, as well. In Fig~\ref{fig:activity_layers}, we assessed the effect of choosing different intermediate layers, among Conv-4, FC-1, and FC-2, on the accuracy-privacy trend of the advanced embedding method. Like what we experienced for gender classification, in activity recognition, considering higher layers as the intermediate results in having more area under the accuracy-privacy curve.

\begin{figure*}[t]
	\centering
	\includegraphics[width = .7\textwidth]{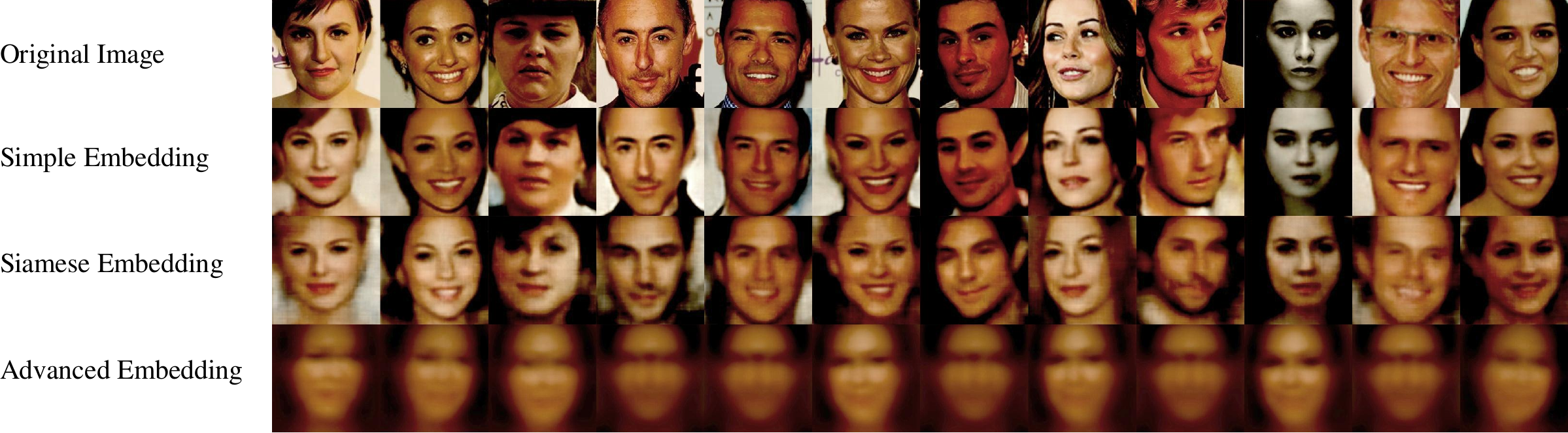} 	
	\caption{The first row shows the original images and the others show the reconstructed ones from exclusive feature. In all reconstructed images, the gender of the individuals is recognized to be the same as the originals. In addition, From simple to advanced embedding, the identity of the individuals is increasingly removed, proving that the \emph{advanced embedding} has the best privacy preservation performance.}
	\label{fig:visualization}
\end{figure*}

\descr{Visualization.} We evaluate the identity privacy of the gender classification task, whose input is an image and thus visualizable, using deep visualization techniques. In order to visualize the output of the feature extractor module, we fed the Alexnet decoder \cite{dosovitskiy2016} with the exclusive feature of the gender classification model for different embeddings. Then we fine-tuned the decoder to reconstruct the input image as much as possible. We can visually verify the identity privacy of the feature extractor by looking at the reconstructed images. The results are illustrated in Fig~\ref{fig:visualization} for different embedding methods. It can be observed that under all of the different embedding methods, genders of all images remain the same as the original ones. However, we can see that for advance embedding, it is harder to distinguish identities from the reconstructed images compared to other embeddings. The original images are mostly recovered in simple embedding. Therefore, just separating layers of a deep network can not guarantee the privacy of users' data. Siamese embedding performs better than simple embedding by distorting the identity, yet advanced embedding provides the best results.

\begin{figure}[t]
	\definecolor{color1}{HTML}{B2B2FF}
	\definecolor{color2}{HTML}{FFB2B2}
	\definecolor{color3}{HTML}{ECD9C6}
	\definecolor{color4}{HTML}{808080}
	\centering
	\begin{subfigure}[t]{.48\columnwidth}
		\centering
		\begin{tikzpicture}
		\begin{axis}[
		tiny,
		width=1.9in,
		boxplot/draw direction=y,
		boxplot={
			%
			draw position={1/3 + floor(\plotnumofactualtype/2) + 1/3*mod(\plotnumofactualtype,2)},
			%
			box extend=0.3,
		},
		xmax=4.2,
		xtick={0,1,...,10},
		x tick label as interval,
		xticklabels={%
			{Conv5-1},%
			{Conv5-2},%
			{Conv5-3},%
			{On-Prem.},%
		},
		ylabel=\textbf{Running Time (ms)},
		x tick label style={
			align=center
		},
		legend entries={Simple, Reduced},
		legend to name={legend},
		name=border
		]
		
		\addplot+ [boxplot, solid, mark=none, fill=color1,draw=blue,postaction={pattern=north east lines,pattern color=blue}]
		table [row sep=\\,y index=0,y expr=\thisrow{data}*0.000001] {
			data
			\\16064916.0\\21466666.0\\21028250.0\\22988459.0\\22198458.0\\22783458.0\\21430458.0\\21215375.0\\21279834.0\\21209500.0\\21418708.0\\23170458.0\\21999625.0\\22357959.0\\21768875.0\\21367167.0\\23722208.0\\22502875.0\\22735667.0\\22826125.0\\22724250.0\\22301667.0\\19523917.0\\22755416.0\\21730000.0\\22252250.0\\22689000.0\\21703833.0\\19367166.0\\19344292.0\\21362792.0\\20937333.0\\21426750.0\\23236500.0\\22896417.0\\22225333.0\\21629417.0\\23134291.0\\22029667.0\\21929292.0\\21868959.0\\23154958.0\\22912459.0\\22729667.0\\21972916.0\\20996792.0\\22390416.0\\21507791.0\\21527125.0\\21878666.0\\22427750.0\\23076167.0\\22843750.0\\21842208.0\\21665791.0\\21359375.0\\22183250.0\\22815916.0\\21415875.0\\20761542.0\\
		};

		\addplot+ [boxplot, solid, mark=none, fill=color2,draw=red,postaction={pattern=north west lines,pattern color=red}]
		table [row sep=\\,y index=0,y expr=\thisrow{data}*0.000001] {
			data\\
			28319250.0\\26215333.0\\23768500.0\\25628000.0\\23993000.0\\24024584.0\\25750250.0\\23460000.0\\24781708.0\\23561084.0\\23781125.0\\24406042.0\\23770416.0\\25230875.0\\24977750.0\\25005459.0\\25688625.0\\25074500.0\\24496334.0\\24255042.0\\23796750.0\\24057834.0\\23753167.0\\24025708.0\\24971542.0\\21328250.0\\21047042.0\\19691583.0\\24726042.0\\24764916.0\\25309083.0\\25760583.0\\24864041.0\\24874125.0\\24267833.0\\26911500.0\\23979250.0\\26452584.0\\26488583.0\\28355375.0\\25278375.0\\23855708.0\\23844959.0\\26718083.0\\25558417.0\\23938458.0\\25799208.0\\24560250.0\\23846125.0\\24515042.0\\25406709.0\\27044375.0\\24787708.0\\25485792.0\\25717583.0\\26824208.0\\25904625.0\\28079416.0\\18165250.0\\24297792.0\\
		};
		
		\addplot+ [boxplot, solid, mark=., fill=color1,draw=blue,postaction={pattern=north east lines,pattern color=blue}]
		table [row sep=\\,y index=0,y expr=\thisrow{data}*0.000001] {
			data\\
			18721916.0\\24718292.0\\22942417.0\\22167833.0\\22749208.0\\22107959.0\\22301750.0\\25065333.0\\24836167.0\\23448375.0\\23254417.0\\25777875.0\\25379125.0\\23191958.0\\22485584.0\\24022459.0\\23059292.0\\23510833.0\\17029792.0\\23534625.0\\22548666.0\\24717708.0\\22023167.0\\23622208.0\\21741709.0\\23208584.0\\21462084.0\\21556458.0\\24786750.0\\24659666.0\\22658000.0\\24274125.0\\22806125.0\\24718708.0\\23551000.0\\25864708.0\\23880583.0\\23465208.0\\22075541.0\\21341750.0\\22088250.0\\23054500.0\\22607709.0\\22688291.0\\21855167.0\\23060000.0\\23222000.0\\23685750.0\\23335042.0\\22588250.0\\21795000.0\\22955625.0\\24787416.0\\24077458.0\\23860250.0\\20784667.0\\27306959.0\\23613916.0\\23094125.0\\23348041.0\\
		};
		
		\addplot+ [boxplot, solid, mark=., fill=color2,draw=red,postaction={pattern=north west lines,pattern color=red}]
		table [row sep=\\,y index=0,y expr=\thisrow{data}*0.000001] {
			data\\
			24625542.0\\23889917.0\\24479000.0\\26976375.0\\24281584.0\\24170416.0\\26973958.0\\24839125.0\\24685417.0\\24124917.0\\23563875.0\\24576750.0\\23853666.0\\24213708.0\\26115333.0\\24106292.0\\24521084.0\\24289666.0\\27254208.0\\25021167.0\\24707083.0\\24005750.0\\23977000.0\\25631917.0\\24051209.0\\25955250.0\\25580083.0\\24350458.0\\23795666.0\\21601792.0\\23618750.0\\23852667.0\\24740334.0\\25898375.0\\26521500.0\\22617458.0\\24272708.0\\24662500.0\\25884542.0\\24748125.0\\23829125.0\\23736959.0\\26936000.0\\25238500.0\\25606167.0\\25219792.0\\24262959.0\\26635875.0\\25226250.0\\25315792.0\\25404208.0\\24210958.0\\24445875.0\\23715000.0\\24737875.0\\26165583.0\\26861250.0\\24581291.0\\23632542.0\\24298917.0\\
		};
		
		\addplot+ [boxplot, solid, mark=., fill=color1,draw=blue,postaction={pattern=north east lines,pattern color=blue}]
		table [row sep=\\,y index=0,y expr=\thisrow{data}*0.000001] {
			data\\
			15038791.0\\21420333.0\\24448958.0\\22433667.0\\22419708.0\\21950208.0\\22847500.0\\24650292.0\\24958875.0\\24021000.0\\22929333.0\\23201875.0\\23089792.0\\23539625.0\\26544959.0\\26723208.0\\25784209.0\\22820292.0\\24302667.0\\23567792.0\\23425916.0\\25952584.0\\23260875.0\\27070042.0\\23894125.0\\22953709.0\\22995875.0\\22651042.0\\26006791.0\\25803417.0\\23948542.0\\25131250.0\\22656958.0\\24644584.0\\23126791.0\\23028167.0\\25958542.0\\23746625.0\\23305083.0\\23814334.0\\23108250.0\\23737125.0\\23496709.0\\24204541.0\\22576000.0\\23323625.0\\23521000.0\\25473708.0\\24589625.0\\25751416.0\\23097250.0\\23591750.0\\22124416.0\\23706750.0\\26969667.0\\22962791.0\\22185792.0\\23891708.0\\25249042.0\\25955417.0\\
		};
		
		\addplot+ [boxplot, solid, mark=., fill=color2,draw=red,postaction={pattern=north west lines,pattern color=red}]
		table [row sep=\\,y index=0,y expr=\thisrow{data}*0.000001] {
			data\\
			21726209.0\\26128917.0\\26782750.0\\28471750.0\\27974500.0\\24913167.0\\20150375.0\\24203000.0\\27430708.0\\24370583.0\\25825875.0\\25530417.0\\19604833.0\\26289042.0\\24854000.0\\26021291.0\\28202000.0\\23601375.0\\24886583.0\\27333625.0\\25927791.0\\26878208.0\\25465417.0\\25749958.0\\24696500.0\\25524375.0\\25627792.0\\25707375.0\\28436583.0\\25036041.0\\26340333.0\\26085208.0\\25317625.0\\26708833.0\\24251250.0\\25465042.0\\26104416.0\\24611584.0\\26520334.0\\28564000.0\\26843542.0\\27029084.0\\24802250.0\\22695208.0\\22378334.0\\22468125.0\\25405625.0\\24408125.0\\25237458.0\\25927750.0\\23699417.0\\25194458.0\\24985458.0\\24082375.0\\24080667.0\\26112292.0\\25373917.0\\24422458.0\\24522125.0\\24763000.0\\
		};
		
		\addplot+ [boxplot, solid, mark=., fill=color4,draw=black]
		table [row sep=\\,y index=0,y expr=\thisrow{data}*0.000001] {
			data\\
			31049500.0\\32202541.0\\35491833.0\\37384125.0\\40083583.0\\34608166.0\\33461875.0\\34060750.0\\34635416.0\\32243583.0\\33483917.0\\32782000.0\\32082708.0\\35688291.0\\33113000.0\\33587334.0\\36943708.0\\33910375.0\\36596500.0\\35621500.0\\32533166.0\\32705041.0\\33325375.0\\37218916.0\\33073792.0\\35340916.0\\33511833.0\\35020667.0\\38695791.0\\33903709.0\\31696583.0\\34179958.0\\34165583.0\\33862166.0\\33858291.0\\40038458.0\\37293709.0\\34787417.0\\32945792.0\\32224042.0\\34305041.0\\34221792.0\\33538417.0\\35790083.0\\34105667.0\\33796625.0\\34219917.0\\33125958.0\\33876000.0\\35033042.0\\34247791.0\\33175125.0\\33883584.0\\34548458.0\\33138417.0\\33846125.0\\34736042.0\\34672958.0\\36050334.0\\31581916.0\\
		};
	

		\end{axis}
		\node[below right] at (border.north west) {\ref{legend}};
		\end{tikzpicture}
		\caption{Gender Classification}
		\vspace*{15pt}
	\end{subfigure}
	\hfill
	\begin{subfigure}[t]{.48\columnwidth}
		\centering
		\begin{tikzpicture}
		\begin{axis}[
		tiny,
		width=1.9in,
		boxplot/draw direction=y,
		boxplot={
			%
			draw position={1/3 + floor(\plotnumofactualtype/2) + 1/3*mod(\plotnumofactualtype,2)},
			%
			box extend=0.3,
		},
		xmax=4.2,ymax=5,ymin=2,
		xtick={0,1,2,...,12},
		x tick label as interval,
		xticklabels={%
			{Conv-4},%
			{FC-1},%
			{FC-2},%
			{On-Prem.},%
		},
		x tick label style={
			align=center
		},
		ylabel=\textbf{Running Time (ms)},
		legend entries={Simple, Reduced},
		legend to name={legend2},
		name=border
		]
		\addplot+ [boxplot, solid, mark=none, fill=color1,draw=blue,postaction={pattern=north east lines,pattern color=blue}]
		table [row sep=\\,y index=0,y expr=\thisrow{data}*0.000001] {
			data
			\\5374792.0\\3198875.0\\3139500.0\\3093917.0\\2963625.0\\3031833.0\\3070250.0\\3302583.0\\3163708.0\\3061667.0\\2903750.0\\3152375.0\\2942625.0\\3035917.0\\2975209.0\\3231834.0\\3112333.0\\3099667.0\\3092959.0\\2917125.0\\3076292.0\\2988791.0\\2994833.0\\3040625.0\\2971209.0\\3122500.0\\2966250.0\\3012167.0\\3016583.0\\3235250.0\\3000250.0\\3180125.0\\3082459.0\\3099917.0\\3114375.0\\3013792.0\\2977125.0\\2998583.0\\3123666.0\\2934500.0\\2930708.0\\3136792.0\\3021708.0\\3218833.0\\3085208.0\\3049834.0\\3098209.0\\3266542.0\\3165000.0\\3165000.0\\3113875.0\\3249958.0\\2993292.0\\3078709.0\\3014375.0\\2581041.0\\2331500.0\\2336250.0\\2113625.0\\2298833.0\\
		};
		\addplot+ [boxplot, solid, mark=none, fill=color2,draw=red,postaction={pattern=north west lines,pattern color=red}]
		table [row sep=\\,y index=0,y expr=\thisrow{data}*0.000001] {
			data
			\\5974375.0\\3208209.0\\2917833.0\\2769666.0\\2472458.0\\2502417.0\\2321917.0\\2259333.0\\2313542.0\\2133166.0\\2178584.0\\2068917.0\\2207959.0\\2330833.0\\1788042.0\\2302750.0\\3043417.0\\3153084.0\\3314834.0\\3167167.0\\2877833.0\\3001375.0\\3034500.0\\3031833.0\\3173083.0\\3179000.0\\3154125.0\\2534750.0\\3203416.0\\3039958.0\\2946000.0\\3060416.0\\3055000.0\\3021042.0\\2950000.0\\3059084.0\\3286875.0\\3367750.0\\3208334.0\\3232625.0\\3144375.0\\3373833.0\\2818500.0\\2958792.0\\2895083.0\\3027000.0\\2910709.0\\2905709.0\\3007959.0\\3189916.0\\3100500.0\\3057958.0\\2997250.0\\3095167.0\\2966667.0\\3039666.0\\2930292.0\\2976667.0\\3042750.0\\3162292.0\\
		};
		\addplot+ [boxplot, solid, mark=., fill=color1,draw=blue,postaction={pattern=north east lines,pattern color=blue}]
		table [row sep=\\,y index=0,y expr=\thisrow{data}*0.000001] {
			data
			\\4277958.0\\3221250.0\\3210834.0\\3093042.0\\3107333.0\\3113584.0\\3067958.0\\3142792.0\\2398125.0\\3081417.0\\3089792.0\\3256583.0\\3010916.0\\3026500.0\\3104166.0\\3114792.0\\3058750.0\\3008125.0\\3043208.0\\2952708.0\\3051583.0\\2925750.0\\2980208.0\\2955541.0\\3020166.0\\2943333.0\\2967375.0\\3054083.0\\3006791.0\\3037208.0\\3266875.0\\3017875.0\\3066667.0\\3028958.0\\3165917.0\\3118875.0\\2967334.0\\3144375.0\\3075166.0\\3096416.0\\3103375.0\\3007083.0\\3019750.0\\3167167.0\\3081875.0\\3004209.0\\3260750.0\\3274750.0\\3326834.0\\3139792.0\\3273000.0\\3142125.0\\3455209.0\\2963083.0\\2837916.0\\3170292.0\\3000583.0\\2934500.0\\2947625.0\\3011042.0\\
		};
		\addplot+ [boxplot, solid, mark=., fill=color2,draw=red,postaction={pattern=north west lines,pattern color=red}]
		table [row sep=\\,y index=0,y expr=\thisrow{data}*0.000001] {
			data
			\\4127792.0\\3140125.0\\3121458.0\\2912750.0\\3016250.0\\2938084.0\\2936209.0\\3120750.0\\2948416.0\\3513083.0\\2911250.0\\2933750.0\\2916875.0\\3002041.0\\2881209.0\\2955750.0\\2880125.0\\2869667.0\\2962458.0\\2997125.0\\2949709.0\\2932000.0\\2920500.0\\2975083.0\\2882041.0\\2982584.0\\2918667.0\\2885833.0\\2935750.0\\2935375.0\\2880292.0\\2967125.0\\2945041.0\\2999167.0\\2916792.0\\2960333.0\\2960708.0\\3203292.0\\2929083.0\\2880459.0\\2914459.0\\2975167.0\\3032750.0\\3029834.0\\3003709.0\\3008250.0\\3047042.0\\2911083.0\\2914917.0\\2953083.0\\2983958.0\\3072042.0\\2955875.0\\3003958.0\\2972667.0\\3103500.0\\3083834.0\\2978833.0\\2921042.0\\2960541.0\\
		};
		\addplot+ [boxplot, solid, mark=., fill=color1,draw=blue,postaction={pattern=north east lines,pattern color=blue}]
		table [row sep=\\,y index=0,y expr=\thisrow{data}*0.000001] {
			data
			\\5044875.0\\2992625.0\\2955542.0\\3042917.0\\2930292.0\\2900917.0\\2905667.0\\3004417.0\\2963334.0\\2854792.0\\3078417.0\\2999541.0\\3012042.0\\2315042.0\\2922458.0\\3054791.0\\3013333.0\\3076334.0\\2960792.0\\3013416.0\\2953208.0\\3272167.0\\3315000.0\\2918791.0\\3202875.0\\3198125.0\\3181125.0\\3034625.0\\2867000.0\\2899583.0\\2846333.0\\3071167.0\\2915000.0\\2898584.0\\2864875.0\\2920250.0\\2865583.0\\3005541.0\\2882625.0\\2930042.0\\3002500.0\\2958625.0\\2939875.0\\2621750.0\\3066583.0\\3033875.0\\2950042.0\\2976125.0\\2878542.0\\2947167.0\\2904375.0\\2983084.0\\2917667.0\\3016000.0\\2998417.0\\2932083.0\\2870084.0\\2911500.0\\2930292.0\\3027125.0\\
		};
		\addplot+ [boxplot, solid, mark=., fill=color2,draw=red,postaction={pattern=north west lines,pattern color=red}]
		table [row sep=\\,y index=0,y expr=\thisrow{data}*0.000001] {
			data
			\\4432667.0\\3271083.0\\3036541.0\\2990708.0\\2968542.0\\2965000.0\\3036917.0\\3005167.0\\2875458.0\\2885208.0\\2985708.0\\2952917.0\\2979625.0\\2998125.0\\2951417.0\\3129625.0\\3050208.0\\2967666.0\\2912041.0\\2965875.0\\3081209.0\\3011000.0\\2936667.0\\2982917.0\\3036875.0\\3085875.0\\3014542.0\\2910250.0\\2999000.0\\2915125.0\\2938917.0\\2985042.0\\2912916.0\\2965125.0\\3017458.0\\3050666.0\\2999583.0\\3155375.0\\3067708.0\\3017291.0\\3065000.0\\3086292.0\\2898375.0\\3065542.0\\2987625.0\\2897167.0\\2978292.0\\2949625.0\\3039542.0\\2960791.0\\2961542.0\\2995333.0\\3155417.0\\2969209.0\\2919333.0\\3093625.0\\2945542.0\\3112500.0\\2935958.0\\2875292.0\\
		};
		\addplot+ [boxplot, solid, mark=., fill=color4,draw=black]
		table [row sep=\\,y index=0,y expr=\thisrow{data}*0.000001] {
			data
			\\5867166.0\\3308166.0\\3098959.0\\3092959.0\\3055958.0\\3243917.0\\3135459.0\\3025042.0\\3034292.0\\3083500.0\\3363958.0\\3078750.0\\3102417.0\\3050000.0\\3101000.0\\3212750.0\\3086584.0\\3175791.0\\3058791.0\\3113125.0\\3459875.0\\3152042.0\\3369916.0\\3269792.0\\3336875.0\\3318708.0\\3317125.0\\3315875.0\\3501833.0\\3303667.0\\3298042.0\\2967208.0\\3030291.0\\3059917.0\\3062834.0\\2902167.0\\3049500.0\\3000000.0\\3260667.0\\3140458.0\\3025458.0\\3100167.0\\3067625.0\\3076833.0\\3019875.0\\2960458.0\\3089625.0\\3004250.0\\3162833.0\\3193125.0\\3108625.0\\3086084.0\\3054250.0\\3068625.0\\3241750.0\\3089334.0\\3087667.0\\3079833.0\\3286666.0\\3263250.0\\
		};
		
		\end{axis}
		\node[below right] at (border.north west) {\ref{legend2}};
		\end{tikzpicture}
		\caption{Activity Recognition}
	\end{subfigure}

	\caption{Running time statistics for different embeddings on mobile device over 60 tests per configuration}
	\label{fig:mobile_time}
\end{figure}
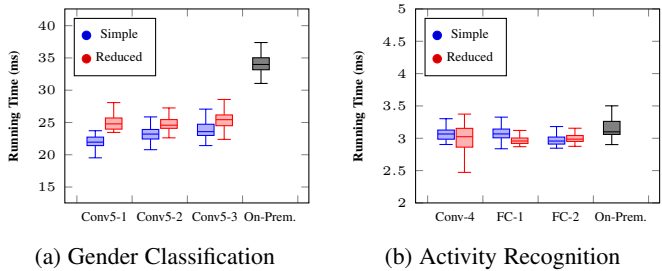

\descr{Mobile Evaluation.}
We evaluated the proposed hybrid framework for both gender classification and activity recognition on a modern handset device, as shown in Table~\ref{tab:device_spec}. In order to assess the time and memory usage of our framework, we evaluated simple and reduced embeddings separately for each of the three intermediate layers, Conv5-1, Conv5-2, and Conv5-3, of VGG-16 architecture and Conv-4, FC-1, and FC-2 of activity recognition model, and compared them with the on-premise solution (full model running on device). Note that we omitted Siamese embedding in mobile evaluations, because the Siamese fine-tuning is done entirely on the cloud, so it does not impact the running time or memory usage of the mobile device. We used Core ML Tools v2.1.0\footnote{https://pypi.org/project/coremltools/} to convert the models into CoreML format that is compatible with iOS platform. Next, we evaluated each model by measuring the running time (Fig~\ref{fig:mobile_time}), model loading time (Fig~\ref{fig:mobile_loading_time}) and model memory usage (Fig~\ref{fig:mobile_memory}) of each of the seven configurations.

\begin{table}[t]
\centering
\caption{Device Specification}
\label{tab:device_spec}
\begin{tabular}{ r l }
	\toprule
	\multicolumn{2}{c}{Apple iPhone Xs} \\
    \midrule
    Memory        & 4GB LPDDR4X RAM \\
	Storage       & 256GB \\
    Chipset       & A12 Bionic chip \\
    CPU           & Hexa-core 64bit \\
    OS            & iOS 12.2 \\
    \bottomrule
\end{tabular}
\end{table}

Most of the variations of trained model architectures under the proposed embedding approach report the same loading time and memory usage performance. As the VGG-16 network of the gender classification is too vast compared to the light activity recognition model, there is a much more significant increase in memory usage, loading time, and running time when loading the on-premise solution in gender classification, proving the efficiency of our framework, especially using heavier deep models. 

\definecolor{color1}{HTML}{B2B2FF}
\definecolor{color2}{HTML}{FFB2B2}
\definecolor{color3}{HTML}{ECD9C6}
\definecolor{color4}{HTML}{808080}

\begin{figure}[t]
	\centering
	\begin{subfigure}[t]{.48\columnwidth}
		\centering
		\begin{tikzpicture}
		\begin{axis}[
		ybar,
		tiny,
		width=1.9in,
		enlarge x limits=0.2,
		bar width=0.05in,
		legend style={at={(0.2,0.99)},legend columns=1, anchor=north},
		symbolic x coords={Conv5-1, Conv5-2, Conv5-3, On-Premise},
		xtick={Conv5-1, Conv5-2, Conv5-3, On-Premise},
		xticklabels={Conv5-1, Conv5-2, Conv5-3, On-Prem.},
		xticklabel style={rotate=0},
		ylabel=\textbf{Loading Time (ms)},
		ymajorgrids,
		y tick label style={
			/pgf/number format/.cd,
			fixed,
			fixed zerofill,
			precision=0,
			/tikz/.cd
		},
		]

		\addplot[fill=color1,draw=blue,postaction={pattern=north east lines,pattern color=blue}] coordinates {(Conv5-1,838.642417) (Conv5-2,790.777042) (Conv5-3,871.369458)};
		
		\addplot[fill=color2,draw=red,postaction={pattern=north west lines,pattern color=red}] coordinates {(Conv5-1,699.012333) (Conv5-2,791.679084) (Conv5-3,871.461375)};
		
		\addplot [fill=color4,draw=black] coordinates {(On-Premise,7107.634417)};
		
		\legend{Simple,Reduced}
		\end{axis}
		\end{tikzpicture}
		\caption{Gender Classification}
		\vspace*{15pt}
	\end{subfigure}
	\hfill
	\begin{subfigure}[t]{.48\columnwidth}
		\centering
		\begin{tikzpicture}
		\begin{axis}[
		ybar,
		tiny,
		width=1.9in,
		enlarge x limits=0.2,	
		bar width=0.05in,
		legend style={at={(0.2,0.99)},
			anchor=north,legend columns=1},
		symbolic x coords={Conv-4, FC-1, FC-2, On-Premise},
		xtick={Conv-4, FC-1, FC-2, On-Premise},
		xticklabels={Conv-4, FC-1, FC-2, On-Prem.},
		ymin=0,ymax=800,
		ylabel=\textbf{Loading Time (ms)},
		ymajorgrids,
		y tick label style={
			/pgf/number format/.cd,
			fixed,
			fixed zerofill,
			precision=0,
			/tikz/.cd
		},
		]
		
		\addplot[fill=color1,draw=blue,postaction={pattern=north east lines,pattern color=blue}] coordinates {(Conv-4,340.390875) (FC-1,341.459084) (FC-2,356.445334)};
		
		\addplot[fill=color2,draw=red,postaction={pattern=north west lines,pattern color=red}] coordinates {(Conv-4,341.459084)(FC-1,337.360833) (FC-2,340.589042)};
		
		\addplot[fill=color4,draw=black] coordinates {(On-Premise,375.614041)};
		
		\legend{Simple,Reduced}
		\end{axis}
		\end{tikzpicture}
		\caption{Activity Recognition}
	\end{subfigure}

	\caption{Loading time for different embeddings on mobile device}
	\label{fig:mobile_loading_time}
\end{figure}
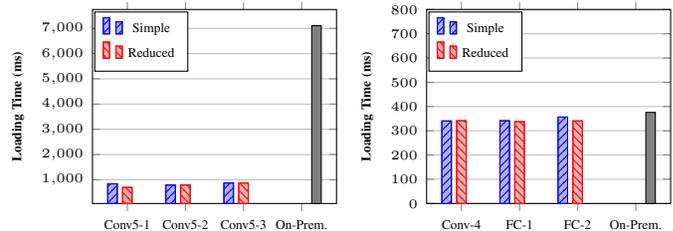


\begin{figure}[t]
	\definecolor{color1}{HTML}{B2B2FF}
	\definecolor{color2}{HTML}{FFB2B2}
	\definecolor{color3}{HTML}{ECD9C6}
	\definecolor{color4}{HTML}{808080}
	\centering
	\begin{subfigure}[t]{.48\columnwidth}
		\centering
		\begin{tikzpicture}
		\begin{axis}[
		ybar,
		tiny,
		width=1.9in,
		enlarge x limits=0.2,
		bar width=0.05in,
		legend style={at={(0.2,0.99)},legend columns=1, anchor=north},
		symbolic x coords={Conv5-1, Conv5-2, Conv5-3, On-Premise},
		xtick={Conv5-1, Conv5-2, Conv5-3, On-Premise},
		xticklabels={Conv5-1, Conv5-2, Conv5-3, On-Prem.},
		xticklabel style={rotate=0},
		ymin=0,ymax=999,
		ylabel=\textbf{Memory Usage (MB)},
		ymajorgrids,
		y tick label style={
			/pgf/number format/.cd,
			fixed,
			fixed zerofill,
			precision=0,
			/tikz/.cd
		},
		]
		
		\addplot[fill=color1,draw=blue,postaction={pattern=north east lines,pattern color=blue}] coordinates {(Conv5-1,108.077184) (Conv5-2,115.884032) (Conv5-3,123.535360)};
		
		\addplot[fill=color2,draw=red,postaction={pattern=north west lines,pattern color=red}] coordinates {(Conv5-1,111.017984) (Conv5-2,117.325824) (Conv5-3,127.893504)};
		
		\addplot [fill=color4,draw=black] coordinates {(On-Premise,986.644480)};
		
		\legend{Simple,Reduced}
		\end{axis}
		\end{tikzpicture}
		\caption{Gender Classification}
		\vspace*{15pt}
	\end{subfigure}
	\hfill
	\begin{subfigure}[t]{.48\columnwidth}
		\centering
		\begin{tikzpicture}
		\begin{axis}[
		ybar,
		tiny,
		width=1.9in,
		enlarge x limits=0.2,
		bar width=0.05in,
		legend style={at={(0.2,0.99)},
			anchor=north,legend columns=1},
		symbolic x coords={Conv-4, FC-1, FC-2, On-Premise},
		xtick={Conv-4, FC-1, FC-2, On-Premise},
		xticklabels={Conv-4, FC-1, FC-2, On-Prem.},
		ymin=10,ymax=90,
		ylabel=\textbf{Memory Usage (MB)},
		ymajorgrids,
		y tick label style={
			/pgf/number format/.cd,
			fixed,
			fixed zerofill,
			precision=0,
			/tikz/.cd
		},
		]
		
		\addplot[fill=color1,draw=blue,postaction={pattern=north east lines,pattern color=blue}] coordinates {(Conv-4,52.035584) (FC-1,53.527680) (FC-2,54.557568)};
		
		\addplot[fill=color2,draw=red,postaction={pattern=north west lines,pattern color=red}] coordinates {(Conv-4,52.576256)(FC-1,54.199424) (FC-2,56.557568)};
		
		\addplot[fill=color4,draw=black] coordinates {(On-Premise,57.492032)};
		
		\legend{Simple,Reduced}
		\end{axis}
		\end{tikzpicture}
		\caption{Activity Recognition}
	\end{subfigure}

	\caption{Memory usage for different embeddings on mobile device}
	\label{fig:mobile_memory}
\end{figure}
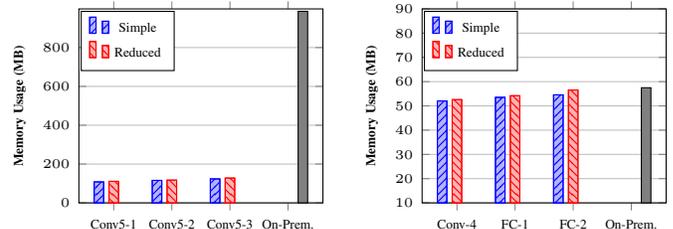


We conclude that our approach is feasible to be implemented in a modern smartphone. By choosing a privacy-complexity trade-off and using different intermediate layers, we were able to significantly reduce the cost when running the model on the mobile device, while at the same time preserving important user information from being uploaded to the cloud.

%% file: related.tex
\section{Related Work}\label{sec:related}

In this section, we first describe the prior works on privacy-preserving machine learning methods and then focus on the case of image analytics. Next, we review the works that have used deep learning on mobile devices.

\subsection{Learning with privacy}

Prior works have approached the problem of privacy in machine learning from a different point of view. Some approaches attempt to remove irrelevant information by increasing the amount of uncertainty, while others try to hide information using cryptographic techniques. Earlier works in this area mainly focus on publicly publishing datasets for machine learning tasks~\cite{agrawal2000, agrawal2001, sweeney2002, iyengar2002}. We categorize these as the \emph{Dataset Publishing} methods. They usually concern about publishing a dataset that consists of high-level features for data mining purposes (e.g., a medical database consisting of patients details), while preserving the individuals' privacy. Solutions such as randomized noise addition~\cite{agrawal2000, agrawal2001} and k-anonymity \cite{sweeney2002} extended by generalization and suppression \cite{lefevre2005, machanavajjhala2007, li2007} are proposed and surveyed in \cite{aggarwal2008}. However, these methods have some major caveats, as they are just appropriate for low-dimensional data due to the curse of dimensionality~\cite{aggarwal2005}. Therefore, they are not suitable for dealing with high-dimensional data, such as multimedia. Furthermore, it is shown that a variety of attacks can make many of these methods unreliable~\cite{aggarwal2008}.

Differential privacy~\cite{dwork2008} is another method that provides an exact way to publish statistics of a database while keeping all individual records of the database private. A learning model trained on a dataset can be considered as its high-level statistics. Therefore, considering the training data privacy while publishing a learning model is another important problem, which we call \emph{Model Sharing}. Many machine learning algorithms were made differentially private in the last decade, surveyed in \cite{ji2014}. Recently, Abadi \etal \cite{abadi2016} proposed differentially private deep models.

A different but related problem to model sharing is the privacy of training data during the training phase. In this problem, which we call it \emph{Training Privacy}, a machine learning model needs to be trained on the data which are distributed among individuals. Centralizing the individual's data may lead to privacy concerns if they contain sensitive information. In this case, the privacy of individuals can be preserved using distributed learning \cite{shokri2015, konevcny2016federated}. In contrast to centralized learning, federated learning methods train a machine learning model in a distributed way by aggregating the parameters. Papernote \etal \cite{papernot2016} introduced a new privacy-preserving framework utilizing differential privacy, which has the state of the art utility-privacy tradeoff. More specific work has been done by Mao \etal that addresses the privacy of individuals when outsourcing the training task of deep convolutional neural networks to an untrusted server \cite{mao2018privacy}.


{All the methods above answer how to train the model such that the training data cannot be revealed from the model. On the contrary, in this paper, we are concerned with the privacy of a service that is being delivered to the users who want to use a model in the cloud, and they are concerned about the privacy of their personal data that is being uploaded. Our method lets the users input the model in the cloud such that their data cannot be stored and used for unauthorized purposes. Hence, we address the privacy issue of users' data from a different perspective compared to dataset publishing, model sharing, and training privacy, including differential privacy and federated learning.

In our scenario, the focus is on the privacy of users' data when they have to upload their data to use an online machine learning based service that is already trained. As a result, neither publishing a dataset, sharing a learned model, nor participating in the training phase are directly relevant to the problem addressed in this paper. Traditional approaches to solving this problem are based on cryptographic methods. In~\cite{avidan2006}, the authors provide a secure protocol for machine learning. In~\cite{gilad2016}, the neural network is held in the cloud, and the input of the network is encrypted in a way that inference becomes applicable to encrypted data, but this approach requires highly complex operations. The neural network should be changed in a complicated manner to enable homomorphic encryption taking 250 seconds on a PC, which makes it impractical in terms of usability on mobile devices. The authors in \cite{rouhani2017, mohassel2017} tried to improve this work by employing a more advanced encryption setting, but they have used simple deep models in their experiments. Recently, Sanyal \etal has further decreased the computational complexity of encryption-based methods using parallelization techniques, but their method still requires more than 100 seconds to be run on a 16-machine cluster \cite{sanyal2018tapas}. Instead of encryption-based methods, an information-theoretic approach is recently introduced in \cite{osia2018dpfe}, where the main focus is on discarding information related to a single user-defined sensitive variable. However, the end-user may not have a complete understanding of what can be inferred from her data to define as sensitive variables. Instead, we enforce the exclusive feature to be specific for the main task of the online service, automatically discarding any other irrelevant information, including those that might be sensitive to users. We use the Siamese architecture to obtain the exclusive feature, which is non-informative for secondary inferences and can be shared with the cloud service.}

 
\subsection{Privacy in image analytics}
A good survey on visual privacy can be found in \cite{padilla2015} that classifies different works into five categories: intervention, blind vision, secure processing, redaction, and data hiding. Our work is similar to de-identification, a subcategory of redaction methods. The goal of these approaches is to perturb the individuals' faces in images in such a way that a face recognition system can not recognize them. A fundamental work in this category is presented in~\cite{newton2005}, which targets privacy issues in video surveillance data. This work aims to publish a transformed dataset, where individuals are not identifiable. They show that using simple image filtering can not guarantee privacy, and suggest \textit{K-same} algorithm based on k-anonymity, aiming to create average face images and replacing them with the original ones. A shortcoming of this work is the lack of protection against future analyses on the dataset. Several works followed this idea and tried to improve it, mainly to publish a dataset of face images. However, they have not considered protecting the privacy of a new face image, which is one of our primary concerns in this paper. Some recent works have tried to transform a face image in a way that it is unrecognizable, while other analytics on the image such as gender classification is possible. Most of the works in this area use visual filters or morphing to make the image unrecognizable~\cite{korshunov2013, rachaud2015}. One of the main issues with prior privacy preservation methods is the lack of a privacy guarantee against new models due to engineering features against specific learning tasks. In most cases, the learning task is not explicitly defined. Moreover, many works ignore the accuracy constraints of the learning task in their privacy preservation methods. 


\subsection{Deep learning on mobile phone}


In the last two years, the implementation and inference ability of deep neural networks on smartphones has experienced a dramatic increase. Using pre-trained deep learning models can increase the accuracy of different sensors. For example, in~\cite{lane2015can}, Lane \etal uses a three layer network that does not overburden the hardware. Complex networks with more layers need more processing power. More complex architectures, such as the 16-layer model (VGG-16) proposed in \cite{simonyan2014very} and the 8-layer model (\textit{VGG-S}) proposed in  \cite{chatfield14}, are implemented on mobile handsets in~\cite{kim2015compression}, and the resource usage, such as time, CPU, and energy overhead, are reported. As most of the state of the art models are large-scale, a full evaluation of all layers on a mobile device results in serious drawbacks in processing time and memory requirements. Some methods are proposed to approximate these complex functions with simpler ones to reduce the cost of inference. Kim \etal~\cite{kim2015compression} tried to compress deep models, while in~\cite{bhattacharya2016}, the authors use sparsification and kernel separation. However, the increase in the efficiency of these methods comes with a decrease in the accuracy of the model. In order to obtain more efficient results, they also implemented the models on GPU. However, the GPU implementation is battery-intensive, making it infeasible for some practical applications that users frequently use or continuously require for long periods \cite{lane2016deepx}. To tackle these problems, Lane~\etal~\cite{lane2016deepx} has implemented a software accelerator called DeepX for large-scale deep neural networks to reduce the required resources while the mobile is doing inference using different kinds of mobile processors. A survey on deep learning on mobile is available in \cite{zhang2018deep}.


%% file: conclusions.tex
\section{Conclusion and next steps}\label{sec:conclusions}

{
	In this paper, we presented a new hybrid framework for efficient privacy-preserving mobile analytics by breaking down a deep neural network into a feature extractor module, which should be deployed on the user's device, and a classifier module, which operates in the cloud. We exploited the properties of DNNs, especially convolutional neural networks, to benefit from their accuracy and layered architecture. In order to protect the data privacy against unauthorized tasks, we used Siamese fine-tuning to prepare an exclusive feature well-suited for the main task, but inappropriate for any other secondary tasks. This is in contrast to ordinary deep neural networks in which the features are generic and may be used for various tasks. Removing the undesired sensitive information from the extracted feature results in the protection of users' privacy. We presented three methods to verify the privacy of the proposed framework and evaluated different embedding methods on various layers of pre-trained state-of-the-art models for gender classification and activity recognition. We demonstrated that our framework could achieve an acceptable trade-off between accuracy and privacy. Furthermore, by implementing the framework on mobile phones, we showed that we could highly decrease the computational complexity on the user side, as well as the communication cost between the user's device and the cloud. 
}

There are many potential future directions for this work trough different case studies. First, the proposed framework is designed to preserve the privacy of users' data when the desired service is a (multi-class) classification task, such as gender recognition or emotion detection. A possible extension to our work is to address this problem for other supervised or unsupervised machine learning problems, such as regression. Second, we would like to provide support for recurrent neural networks to handle time-series input data, such as speech or video. {Also, the proposed solution for balancing the trade-off between accuracy, privacy, and complexity is naive, and the hyper-parameters can be optimized better.} Finally, our current framework is designed for learning inferences in the test phase. In ongoing work, we plan to extend our method by designing a framework for \emph{Machine Learning as a Service}, where the users could share their data in a privacy-preserving manner, to train a new learning model.